\documentclass[]{spie}  

\usepackage{amsmath,amsfonts,amssymb}
\usepackage{graphicx}
\usepackage[colorlinks=true, allcolors=blue]{hyperref}
\usepackage{mathtools}

\usepackage{caption}
\usepackage{subcaption}
\usepackage[section]{placeins}
\usepackage{braket}
\usepackage{bbm}

\usepackage{algorithm}
\usepackage{algorithmic}

\title{Using Deep Image Prior to Assist Variational Selective Segmentation Deep Learning Algorithms}

\author{Liam Burrows\supit{a}, Ke Chen\supit{a} and Francesco Torella\supit{b}
\skiplinehalf
\supit{a}Department of Mathematical Sciences and Centre for Mathematical Imaging Techniques, University of Liverpool, United Kingdom; \\
\supit{b}Liverpool Vascular \& Endovascular Service, Liverpool University Hospitals, Liverpool L7 8XP, United Kingdom
}

\authorinfo{Further author information: (Send correspondence to Ke Chen.)\\Ke Chen.: E-mail: k.chen@liv.ac.uk,}

\begin{document} 
  \maketitle 

\begin{abstract}
Variational segmentation algorithms require a prior imposed in the form of a regularisation term to enforce smoothness of the solution. Recently, it was shown in the Deep Image Prior work that the explicit regularisation in a model can be removed and replaced by the implicit regularisation captured by the architecture of a neural network. The Deep Image Prior approach is competitive, but is only tailored to one specific image and does not allow us to predict future images. We propose to  incorporate the ideas from Deep Image Prior into a more traditional learning algorithm to allow us to use the implicit regularisation offered by the Deep Image Prior, but still be able to predict future images.
\end{abstract}

\section{INTRODUCTION}

Image segmentation has been a well studied topic over the last few decades. The variational approach has been particularly successful in tackling a wide range of problems \cite{mumford1989optimal,chan2001active,cai2013two} without the need for large training sets or time consuming manual labels. A less studied variant is selective variational image segmentation \cite{roberts2019chan,roberts2019convex,spencer2015convex}, which is the task of segmenting a particular object or objects, usually indicated by a set of marker points $\mathcal{M}$ input by the user. These marker points are typically placed inside the region of interest, and geometric constraints can be designed to penalise segmenting regions away from the intended object. 

Deep learning methods for image segmentation \cite{ronneberger2015u,chen2017deeplab,skourt2018lung,li2020joint,cheng2020learning,la2020automated} have become very popular over the last few years due to an increase in computing power and more data availability. A deep learning model tailored to a specific task is the gold standard for image segmentation currently. This is however dependent on a large dataset with ground truth labels, which typically can be rather expensive and/or time consuming to acquire. 

The intersection of variational methods and deep learning methods can lead to a powerful tool to base a variational model in a deep learning framework, potentially reducing the need for many expensive ground truth labels in the form of semi-supervised or unsupervised algorithms, while still leveraging the impressive performance of learning methods.
Such semi-supervised works often combine deep learning methods with variational methods \cite{chen2019multi,burrows2020new,sedai2017semi,kalluri2019universal}.

Another approach that can be used to tackle variational models using deep learning methods was proposed recently by Ulyanov et al. \cite{ulyanov2018deep} in their work on the Deep Image Prior (DIP). Their method uses an untrained neural network to solve an inverse problem for a single image. No training data is required, and surprisingly the DIP approach is competitive with state of the art methods. They propose to remove the explicit regularisation term typically found in a typical inverse problem, and instead rely on the implicit regularisation offered by the architecture of a neural network. 

A major limitation of the Deep Image Prior is that the network trained is tailored to the specific image it is trained on; therefore it is slow and too computationally expensive to be used in practical applications. Prediction to other images of a similar class is not possible as in traditional deep learning methods, as the network is tailored to a specific image only. Furthermore early stopping is required to preserve the regularisation effect. Some work has been done to add more regularisation into the DIP method \cite{liu2019image,mataev2019deepred,van2018compressed} to reduce the need for early stopping. 

In \cite{van2018compressed} the authors used a learned regularisation term in the loss function, where they trained a standard DIP network on a small number of examples on networks individually, saved the weights of each of the networks, and used the prior information of the saved weights in the loss function as a regularisation term in order to train a new test image. Using this learned regularisation term reduces the reconstruction error when used to train new images, however it is still a disadvantage that for every new example a new network has to be trained.

In this work, we aim to use the idea of the Deep Image Prior method to implement a variational model without explicit regularisation in a deep learning model to achieve selective segmentation. We propose to utilise a network architecture that incorporates DIP, but allows for prediction of images in the same class without further training. Our network is provided with image and geometric information to aid the prediction.

\section{Related works}

An inverse problem in imaging is typically formulated as the following: 
\begin{align*}
u^* = \arg \min_{u} \Big\{\mathcal{E}(u;f) + \mathcal{R}(u)\Big\},
\end{align*}
where $f$ is an observed image and $u$ is the desired solution. In this case, $\mathcal{E}(u;f)$ is some data fidelity term relating to a specific task and $\mathcal{R}(u)$ is a regularisation term imposing some kind of prior information. 
It was shown by Ulyanov et al. \cite{ulyanov2018deep} that the explicit regularisation term could be dropped by parametrising the problem using a neural network. The idea behind the Deep Image Prior is that we can write an equivalent problem as the following:
\begin{align*}
\Theta^* = \arg \min_{\Theta} \mathcal{E}(\varphi_{\Theta}(z);f), \qquad u^* = \varphi_{\Theta^*} (z),
\end{align*}
where $\varphi_{\Theta}$ is a neural network parametrised by weights $\Theta$ and $z$ is random noise as input. It was found that the implicit regularisation offered by the architecture of the network was enough to provide a smoothing effect. The implication is that convolutional models tend to fit to smooth signals before fitting to noise; early stopping is key in the DIP method, to prevent overfitting to noise.

The selective segmentation variational model which we will base our proposed approach is the model by Roberts and Spencer \cite{roberts2019chan}. Suppose we have an image $f$ defined on an image domain $\Omega$, their model is the following:
\begin{align}
u^* = \arg \min_{u} \mu \int_{\Omega} g | \nabla u | \, d \mathbf{x} + \lambda \int_{\Omega} \Phi(f,c_1) u \, d \mathbf{x} + \theta \int_{\Omega} \mathcal{D}_G u \, d \mathbf{x},
\label{eq:RobertsSpencer}
\end{align}
where $\Phi$ is a reformulated Chan-Vese term introduced in \cite{roberts2019chan}, $g = g(|\nabla f |) = \frac{1}{1+\iota | \nabla f |^2}$ is an edge detector, $c_1$ is the average intensity of $f$ inside the region of interest and $\mathcal{D}_G$ is the geodesic distance from $\mathcal{M}$ defined in \cite{roberts2019convex}. The binary segmentation result $\Sigma$ is found by thresholding the minimiser, i.e. $\Sigma = \{ \mathbf{x} \in \Omega : u^*(\mathbf{x}) > \gamma \},$ where we usually set $\gamma = 0.5.$ In \cite{burrows2020new}, the authors implemented a similar selective segmentation model to \eqref{eq:RobertsSpencer} in a variational framework.

In variational segmentation models, Total Variation (TV) \cite{rudin1992nonlinear} is a popular choice for regularisation. In \cite{zhu2013image}, Zhu et al. investigated using Euler elastica for regularisation in place of total variation for segmentation. While Euler elastica proved to be an improvement over total variation, minimising it is computationally difficult as the resulting equations are highly non-linear. Among other methods of regularisation, Total Generalised Variation (TGV) \cite{bredies2010total} has also been used as a regulariser in segmentation \cite{duan2016new}. Our proposed approach investigates the effect of removing such explicit regularisation terms, using the Deep Image Prior method.

\section{Proposed approach}

In this section we introduce our proposed approach, which combines the Deep Image Prior approach with a variational model, allowing us to remove the explicit regularisation while preserving the ability to predict other images of the same class.
Our idea is to make use of the Deep Image Prior's ability to extract and learn from deep features in the image providing regularisation in the solution, to remove the need for explicit regularisation in the variational model. Combining it with a more traditional deep learning model allows us to use the combination to predict further images while utilising the regularisation from DIP.

\begin{figure}[!htb]
    \includegraphics[width=\textwidth]{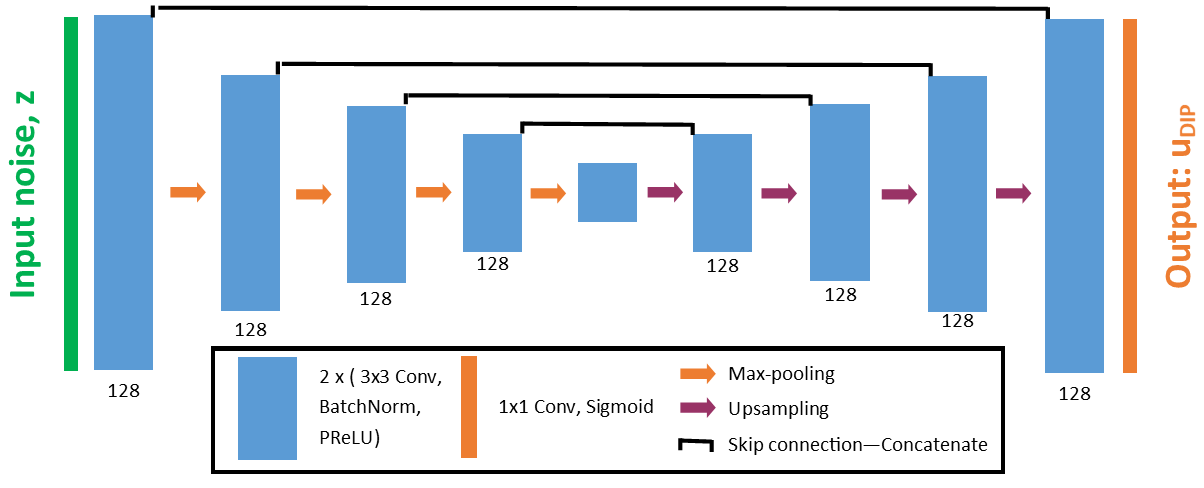}
  \caption{ The architecture of the Deep Image Prior network (DIP Net) we use. A fixed noise vector $z$ is used as input, and output is the segmentation result, $u_{DIP}$.}
  \label{fig:DIPNet}
\end{figure}

Let us denote $f$ as an image ($f \in \mathbb{R}^{N \times M})$, $z$ as a fixed noise vector ( $z \in \mathbb{R}^{N \times M \times 32} \sim U(0,\frac{1}{10}))$ and $\mathcal{G}$ as geometric information associated to the image $f$. For this paper, we consider $\mathcal{G}$ to be either the binary mask formed by the marker set $\mathcal{M}$, or the geodesic distance $\mathcal{D}_G$. Our proposed approach is to have two networks: A variational method like network (VM Net) from \cite{burrows2020new} which we denote as $\psi_{\Theta_1}$, and a deep image prior network which we denote as $\varphi_{\Theta_2}$ (architecture can be found in Figure \ref{fig:DIPNet}), where $\Theta_1$ and $\Theta_2$ are the weights that parametrise each network respectively. Each network produces a segmentation output, $u_{VM}$ and $u_{DIP}$ i.e. $u_{VM} = \psi_{\Theta_1}(f,\mathcal{G})$ and $u_{DIP} = \varphi_{\Theta_2}(z)$. Note that the VM network takes in image information as input $(f$ and $\mathcal{G}$) where the DIP network takes noise as input ($z$) as in the original \cite{ulyanov2018deep}.

\begin{figure}[!htb]
    \includegraphics[width=\textwidth]{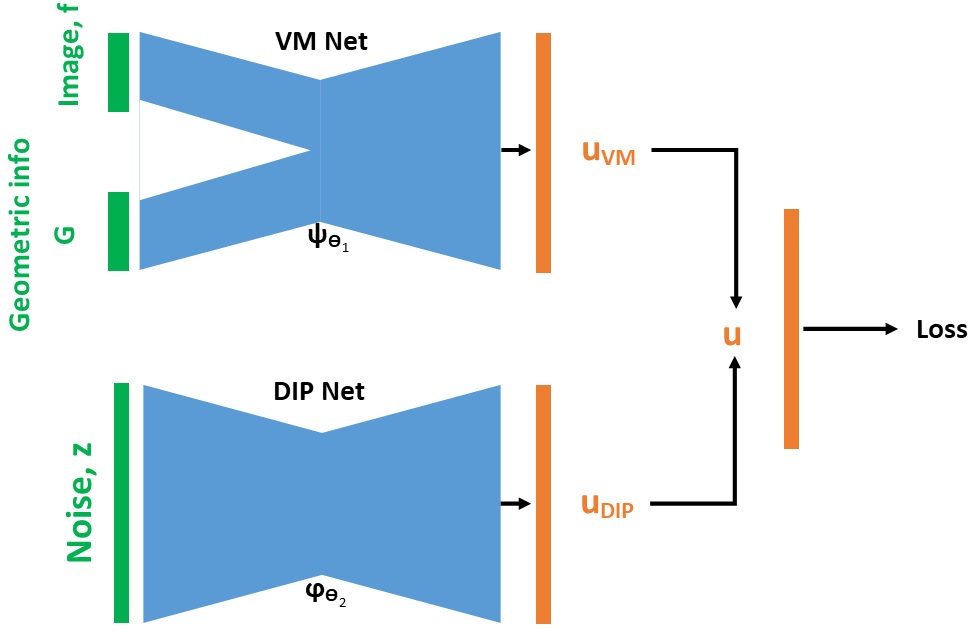}
  \caption{Our overall proposed workflow: Both networks produce a segmentation result, and the two results are combined using a Hadamard product. }
  \label{fig:workflow}
\end{figure}

Overall, we propose to merge these two networks as shown in Figure \ref{fig:workflow}. The idea is to combine each segmentation output, $u_{VM}$ and $u_{DIP}$ with a Hadamard product at the end to obtain $u$. This final $u$ is used in the loss function.

\paragraph{Loss function:}
Our loss function is formed by taking the variational model \eqref{eq:RobertsSpencer} and removing the explicit regularisation term, inspired by the Deep Image Prior \cite{ulyanov2018deep}. Moreover we include an additional term in order to enforce similarity between the outputs of the two networks. The loss function is given as follows:
\begin{align}
\mathcal{L}(\Theta_1, \Theta_2) =  \sum_{j=1}^N & \lambda \int_{\Omega}  \Phi(f^{(j)},c_1^{(j)}) u^{(j)} d \mathbf{x} + \theta \int_{\Omega} \mathcal{D}_G^{(j)} u^{(j)} d \mathbf{x} \nonumber \\
& + \frac{1}{2} \int_{\Omega} ( u^{(j)}_{DIP} - u^{(j)}_{VM} )^2 d \mathbf{x},
\label{eq:proposedloss}
\end{align}
where $N$ is the total number of training images, $u^{(j)}$ denotes the output of the combined network of the $j'th$ training image, $u_{VM}^{(j)} = \psi_{\Theta_1}(f^{(j)},\mathcal{G}^{(j)})$ and $u_{DIP}^{(j)} = \varphi_{\Theta_2}(z^{(j)})$. Therefore, $u_{VM} = u_{VM}(\Theta_1)$ and $u_{DIP} = u_{DIP}(\Theta_2),$ and so $u = u(\Theta_1, \Theta_2)$. Moreover, $\mathcal{D}_G^{(j)}$ is the geodesic distance calculated using the marker points $\mathcal{M}^{(j)}$ on the $j'th$ image, and $c_1^{(j)}$ is fixed as the average intensity inside the polygon formed by $\mathcal{M}^{(j)}$. Training a network with this loss function does not require any ground truth labels, and can be used to predict future unseen images unlike the original DIP work \cite{ulyanov2018deep} which fundamentally cannot predict a new image. 

We do not require an explicit regularisation term as we utilise the implicit regularisation offered by the DIP. During the training stage, we rely on the regularisation effect offered by the DIP net to influence the weights of the VM net, which is why the third term in \eqref{eq:proposedloss} is present. We note that, as in the original work \cite{ulyanov2018deep}, early stopping of our network is still required to preserve the regularisation effect. In our experiments, we find only a very small number of $N$ is required (we use $N= 2$). As the principle of the Deep Image Prior is to underfit an overparametrised network to maintain a regularisation effect, we find that large values of $N$ do not make sense. It is important that we therefore supplement the network with additional information, such as geometric information based on user input like the geodesic distance.

We stress that for each training image $f^{(j)}$ there is a unique fixed noise vector $z^{(j)}$ associated to that training image. At test stage, we only consider the output of the VM net for segmentation. That is to say, we fix the weights of the VM net and the DIP net is no longer needed as the regularisation effect from the DIP net has already been transferred to the VM net during training. Therefore to predict a test image we only require the VM net and the inputs $f$ and $\mathcal{G}$ for the test image, $u = \psi_{\Theta_1}(f,\mathcal{G})$.

\section{Methods}
The proposed method was implemented in Tensorflow 2.3.0 using Keras backend. For our networks, we used the Adam optimiser with a learning rate of $0.001$. As inputs, for the VM Net we use the image $f$ and geometric information $\mathcal{G}$ as inputs. When $\mathcal{G}$ is $\mathcal{M}$, this is simply a binary image, where the region inside the polygon formed by the markers is $1$, and $0$ outside. When $\mathcal{G}$ is $\mathcal{D}_G$, we calculate the geodesic distance from $\mathcal{M}$ by solving the Eikonal equation  detailed in \cite{roberts2019convex}. For the DIP net, we use input $z \in \mathbb{R}^{N \times M \times 32}  \sim U(0,\frac{1}{10})$, however as in \cite{ulyanov2018deep} we vary the noise input at each epoch with noise drawn from a normal distribution. Therefore at each training epoch, the input to the DIP net is $z + \hat{z}$, where $\hat{z} \sim \mathcal{N}(0,\frac{1}{100}),$ where $\hat{z}$ is different each epoch.

We will consider four deep learning methods as a comparison, where two use old ideas, and two use the proposed network and loss function. To introduce the four methods, we introduce a baseline loss function that includes explicit regularisation:
\begin{align}
\mathcal{L}(\Theta) = \sum_{j=1}^N \quad  \mu \int_{\Omega} g^{(j)} | \nabla u ^{(j)} | \, d \mathbf{x} + \lambda \int_{\Omega}  \Phi(f^{(j)},c_1^{(j)}) u^{(j)} \, d \mathbf{x}
+ \theta \int_{\Omega} \mathcal{D}_G^{(j)} u^{(j)} \, d \mathbf{x},
\label{eq:DLLossBaseline}
\end{align}
where $u=u(\Theta)$ is the output of the network, $g^{(j)}$ is an edge detector associated to the $j'th$ image, and $c_1^{(j)}$ and $\mathcal{D}_G^{(j)}$ are defined as in \eqref{eq:proposedloss}.

The four deep learning methods are:

\begin{itemize}
\item \textbf{M1}: The VM Net only, where $\mathcal{G}$ = $\mathcal{M}$ with loss function \eqref{eq:DLLossBaseline}. This method is most similar to the one proposed in \cite{burrows2020new}.

\item \textbf{M2}: The VM Net only, where $\mathcal{G} = \mathcal{M}$ with loss function \eqref{eq:DLLossBaseline} where $\mu = 0$. I.e. without an explicit regularisation term.

\item \textbf{M3}: The proposed network detailed in Figure \ref{fig:workflow}, where $\mathcal{G} = \mathcal{M}$ with proposed loss function \eqref{eq:proposedloss}.

\item \textbf{M4}: The proposed network detailed in Figure \ref{fig:workflow}, where $\mathcal{G} = \mathcal{D}_G$ with proposed loss function \eqref{eq:proposedloss}.
\end{itemize}

The two methods involving our proposed approach are \textbf{M3} and \textbf{M4}. These four methods are trained using training images and results on testing images are predicted after optimisation, with the weights of the network fixed. As a further comparison, we compare our work with a further three method whose methods require optimisation to be performed for each new image. We use a general form of the model \eqref{eq:RobertsSpencer}:
\begin{equation}
    u^* = \arg \min_u \mathcal{E}(u;f) + \mathcal{R}(u), 
    \label{eq:RS-General}
\end{equation}
where $\mathcal{E}(u;f) = \lambda \int_{\Omega} \Phi(f,c_1) u \, d\mathbf{x} + \theta \int_{\Omega} \mathcal{D}_G u \, d \mathbf{x}.$ The three further methods we compare with are:
\begin{itemize}
    \item \textbf{TV-model:} The model \eqref{eq:RS-General} where $\mathcal{R}$ is the total variation regulariser \cite{rudin1992nonlinear}, i.e. $$\mathcal{R}(u) = \int_{\Omega} | \nabla u | \, d \mathbf{x}.$$
    We solve this model numerically using ADMM.
    
    \item \textbf{Elastica-model:} The model \eqref{eq:RS-General} where $\mathcal{R}$ is the Euler-Elastica regulariser \cite{zhu2013image}, i.e. $$\mathcal{R}(u) = \int_{\Omega} \Big( \alpha + \beta \Big( \nabla \cdot \Big(\frac{\nabla u}{| \nabla u |} \Big)^2 \Big) \Big) | \nabla u| \, d \mathbf{x}.$$
    We solve this model numerically using ADMM.
    
    \item \textbf{DIP-like-model:} The model \eqref{eq:RS-General} optimised as in the original DIP work \cite{ulyanov2018deep}, i.e. the eplxicit regularisation term $\mathcal{R}(u)$ is not needed, and we parametrise the problem using an untrained network $\varphi_{\Theta}$, and the weights are optimised by minimising 
    $$ \min_{\Theta} \mathcal{E}(\varphi_{\Theta};f),$$
    where early stopping is implemented to preserve a regularisation effect. We solve this model using Tensorflow using the conventional Deep Image Prior \cite{ulyanov2018deep} setup.
\end{itemize}

\begin{figure}[!htb]
\centering
 \begin{subfigure}[b]{0.16\textwidth}
    \includegraphics[width=\textwidth]{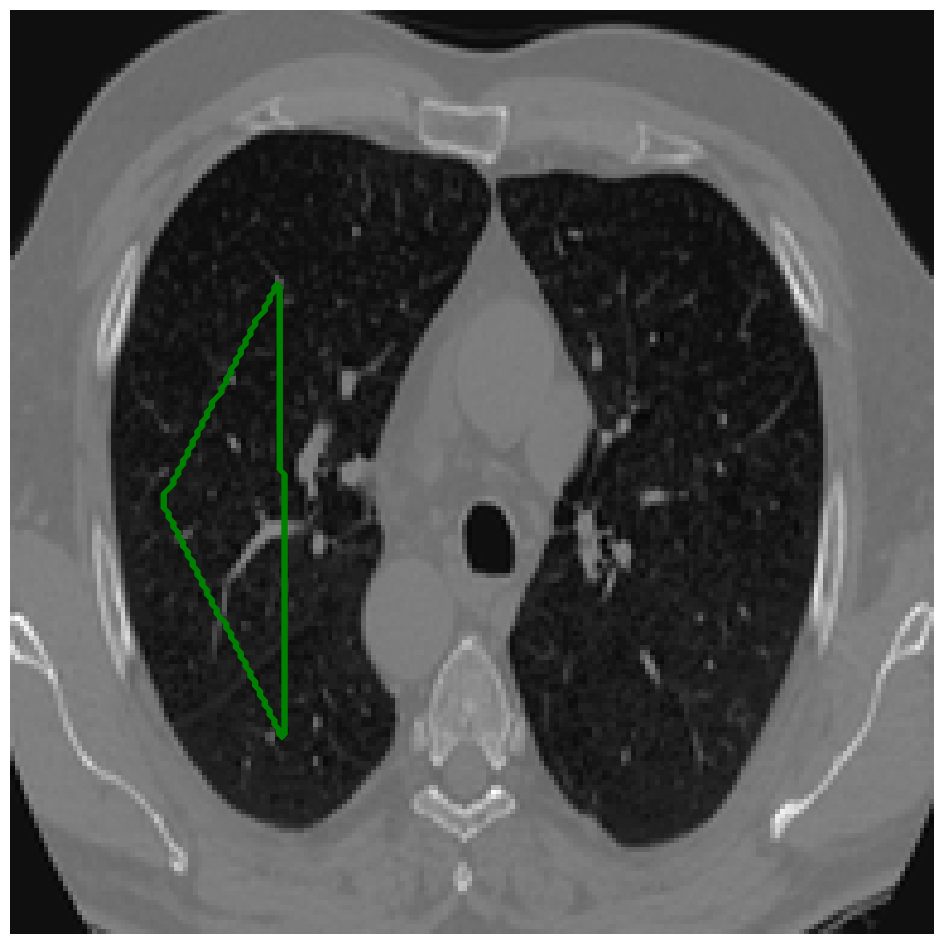}
    \caption{Image}
 \end{subfigure}
  \begin{subfigure}[b]{0.16\textwidth}
    \includegraphics[width=\textwidth]{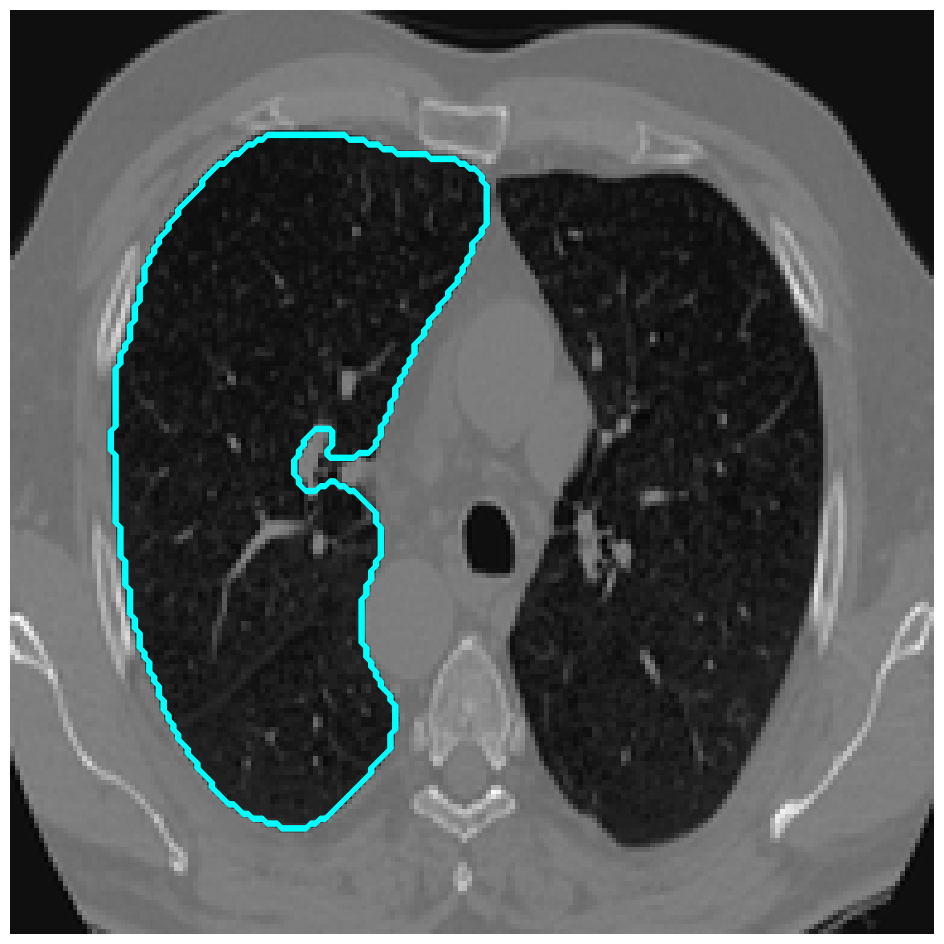}
    \caption{GT}
 \end{subfigure}
  \begin{subfigure}[b]{0.16\textwidth}
    \includegraphics[width=\textwidth]{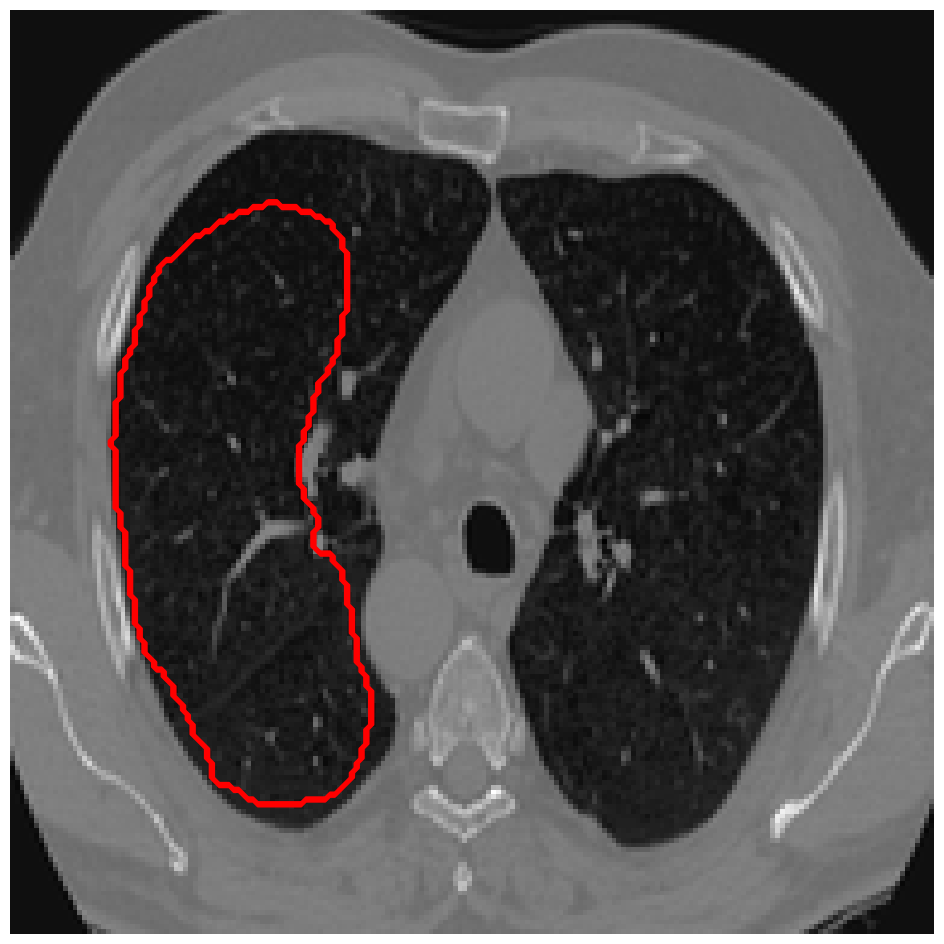}
    \caption{\textbf{M1}}
 \end{subfigure} 
   \begin{subfigure}[b]{0.16\textwidth}
    \includegraphics[width=\textwidth]{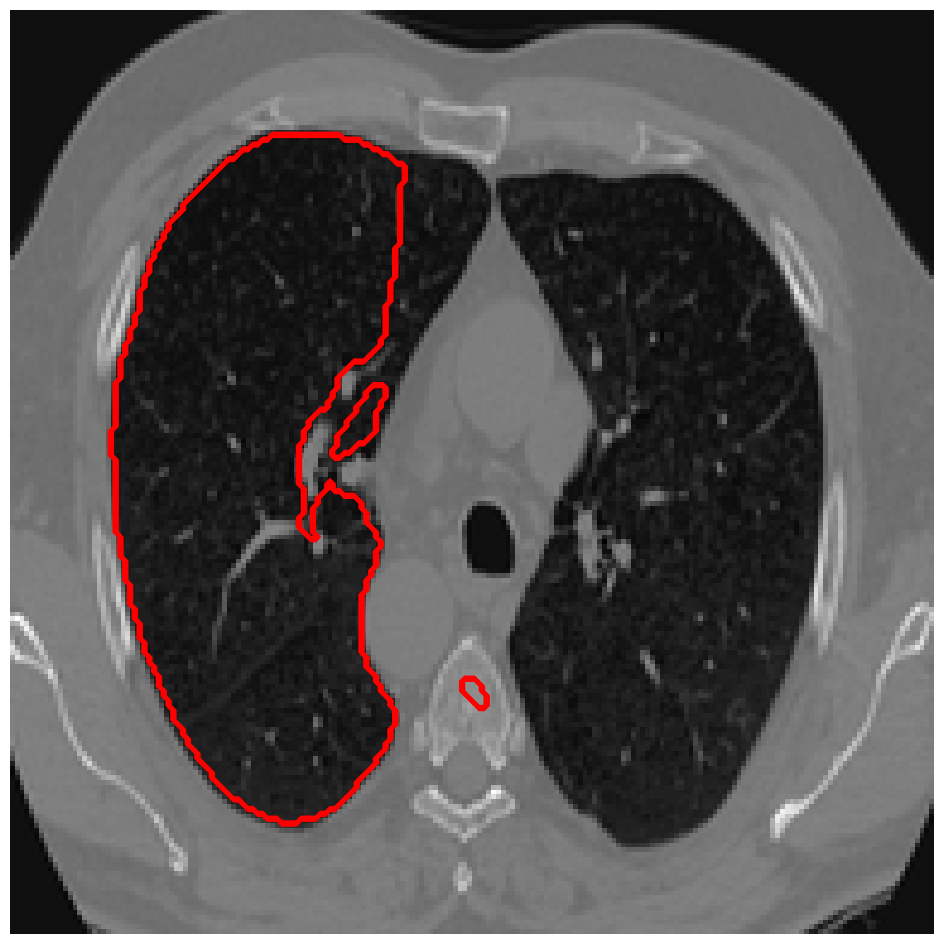}
    \caption{\textbf{M2}}
 \end{subfigure} \\
   \begin{subfigure}[b]{0.16\textwidth}
    \includegraphics[width=\textwidth]{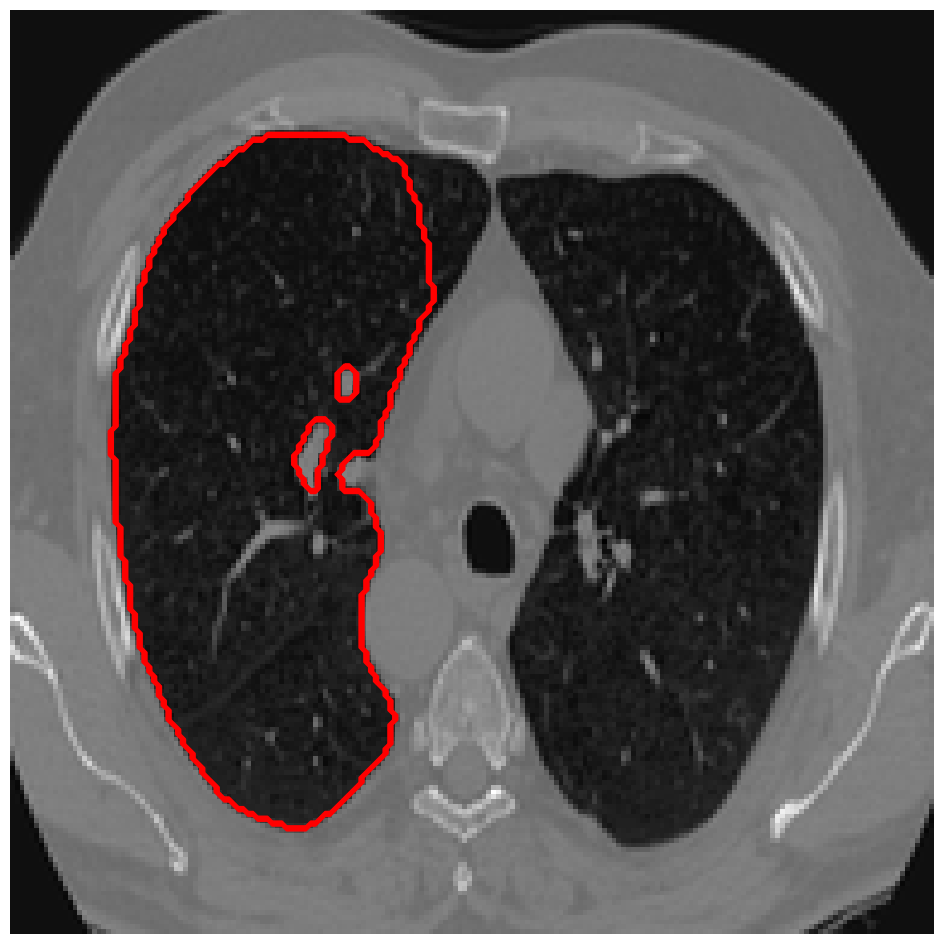}
    \caption{\textbf{M3}}
 \end{subfigure}
   \begin{subfigure}[b]{0.16\textwidth}
    \includegraphics[width=\textwidth]{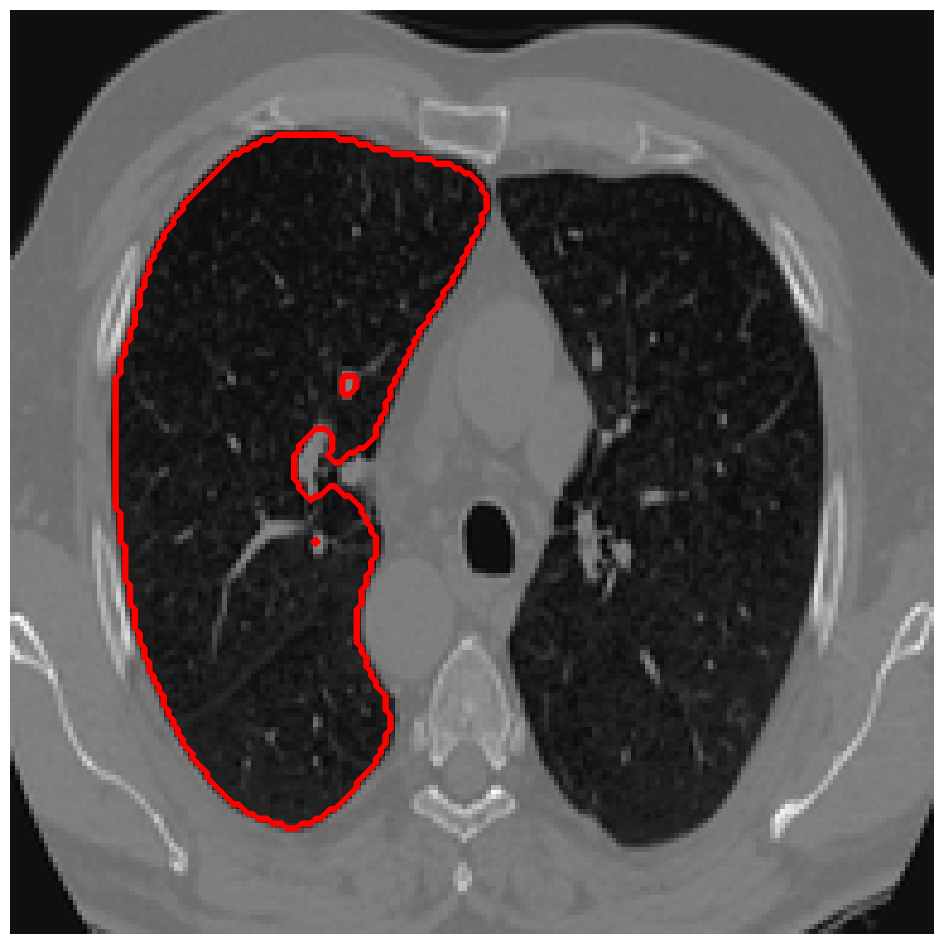}
    \caption{\textbf{M4}}
 \end{subfigure} 
 \begin{subfigure}[b]{0.16\textwidth}
    \includegraphics[width=\textwidth]{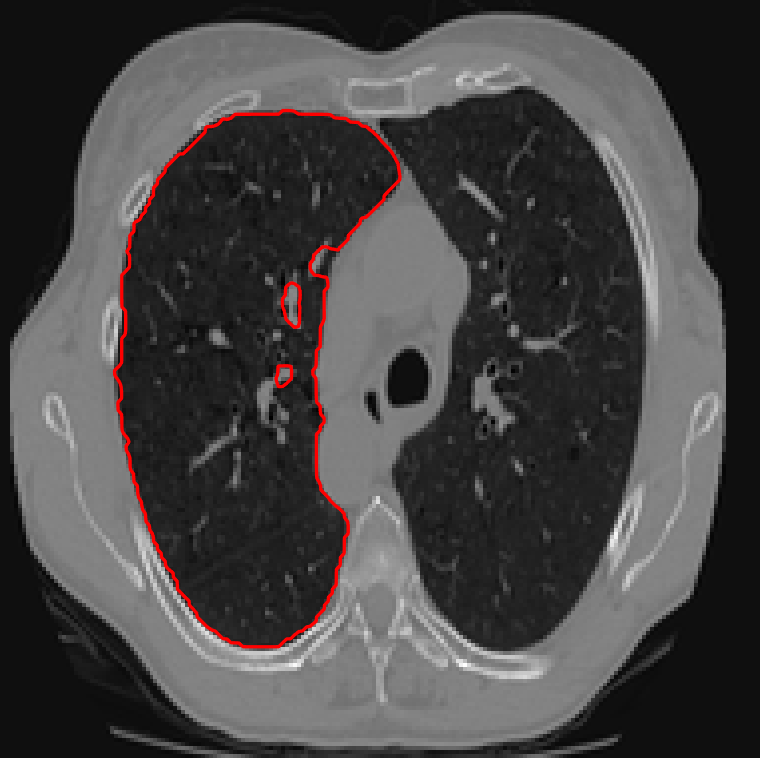}
    \caption{\textbf{TV-model}}
 \end{subfigure} 
 \begin{subfigure}[b]{0.16\textwidth}
    \includegraphics[width=\textwidth]{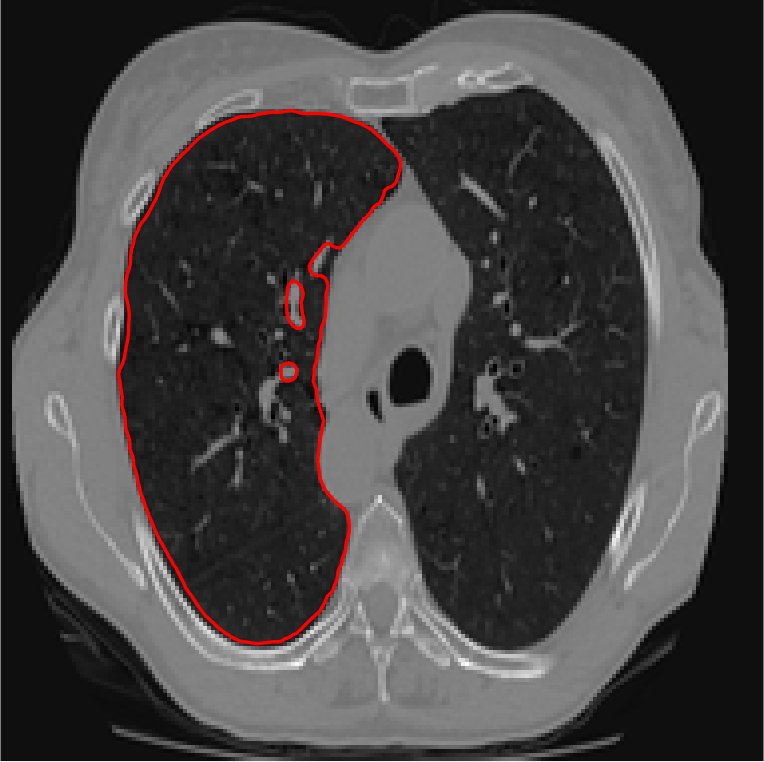}
    \caption{\scriptsize{\textbf{Elastica-model}}}
 \end{subfigure} 
 \begin{subfigure}[b]{0.16\textwidth}
    \includegraphics[width=\textwidth]{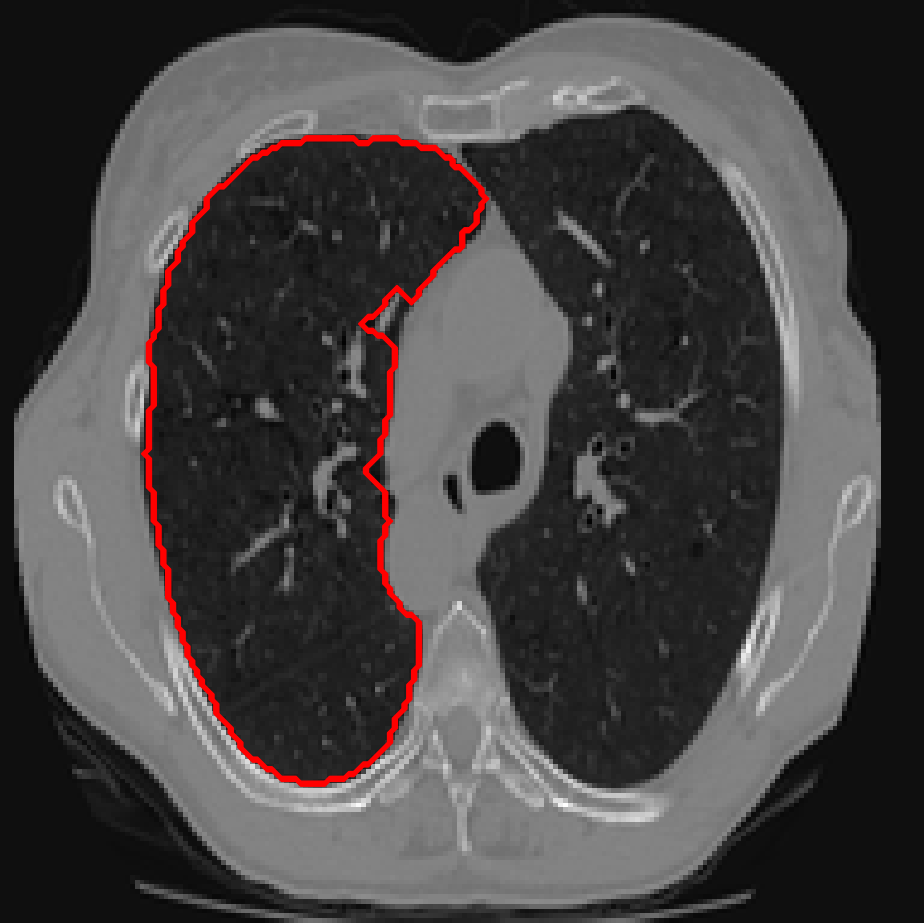}
    \caption{\scriptsize{\textbf{DIP-like-model}}}
 \end{subfigure} 
 \caption{A sample result on the Lung data. We display the input image with the user input $\mathcal{M}$, the ground truth (GT) and results from the four methods. Moreover, we show comparisons with the model \eqref{eq:RS-General} solved in  a variational framework with both Total Variation (TV) and Euler Elastica as explicit regularisation, as well as a comparison with the model solved in a Deep Image Prior framework.}
 \label{fig:Lung1}
\end{figure}  
\begin{figure}[!htb]
\centering
 \begin{subfigure}[b]{0.16\textwidth}
    \includegraphics[width=\textwidth]{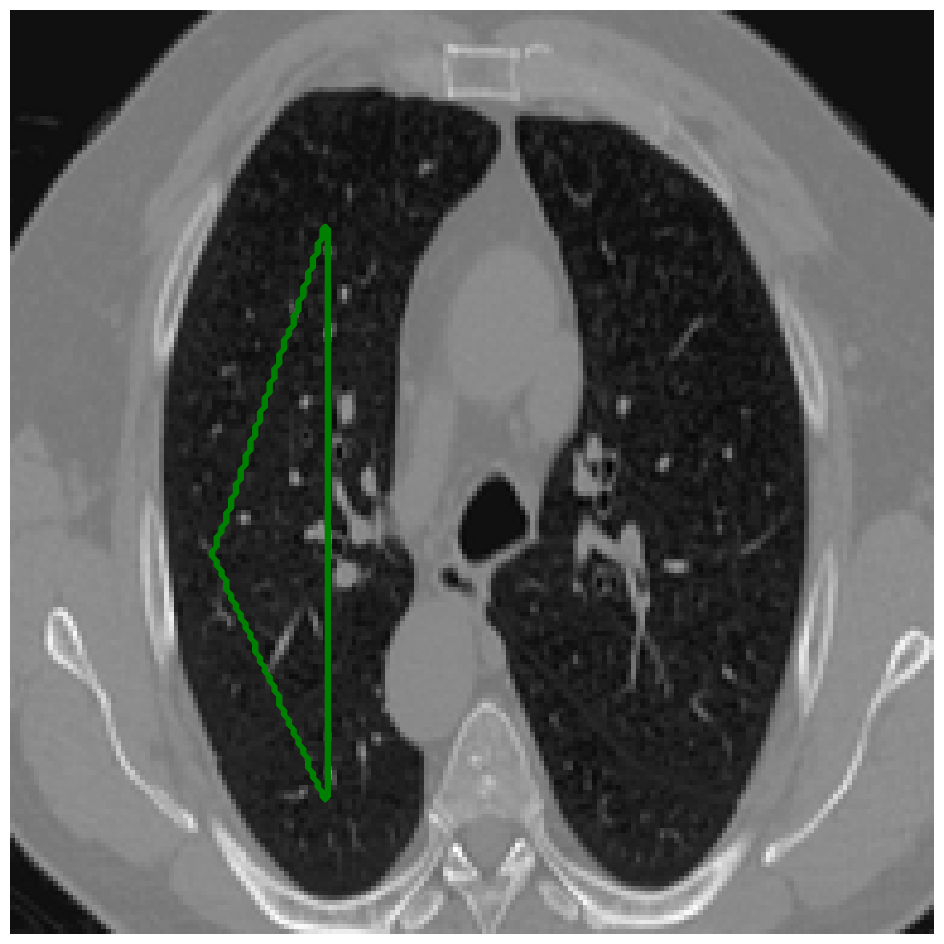}
    \caption{Image}
 \end{subfigure}
  \begin{subfigure}[b]{0.16\textwidth}
    \includegraphics[width=\textwidth]{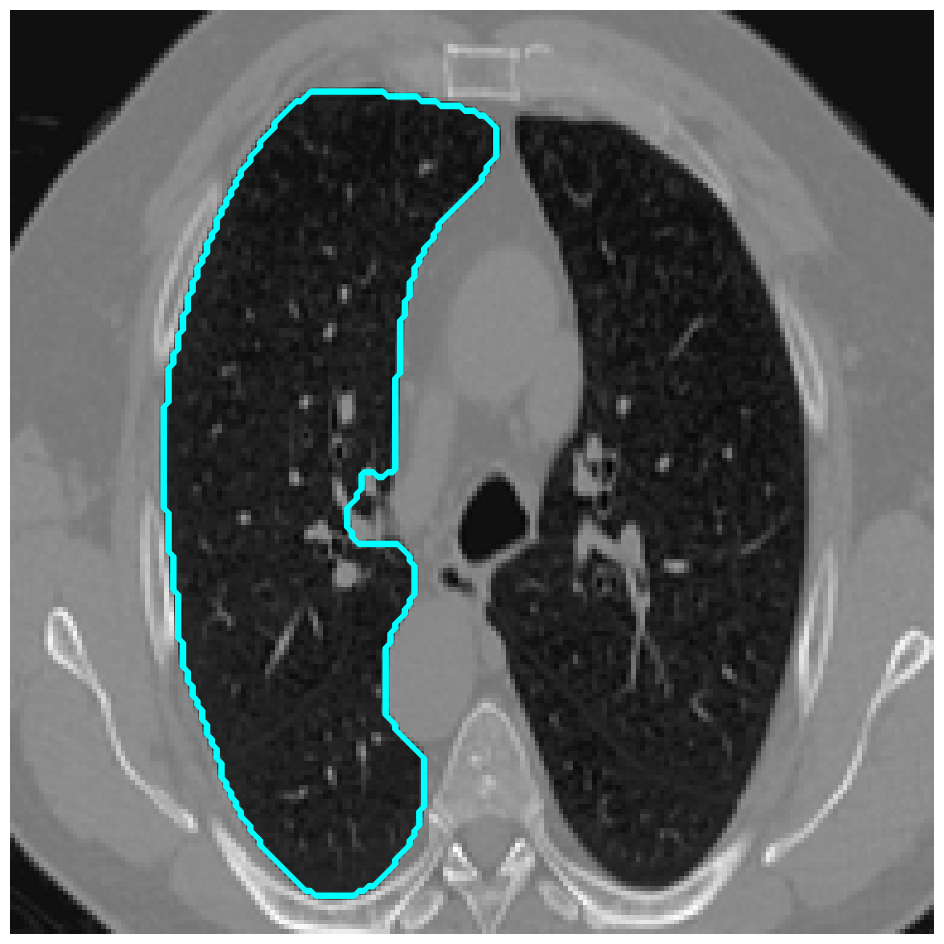}
    \caption{GT}
 \end{subfigure}
  \begin{subfigure}[b]{0.16\textwidth}
    \includegraphics[width=\textwidth]{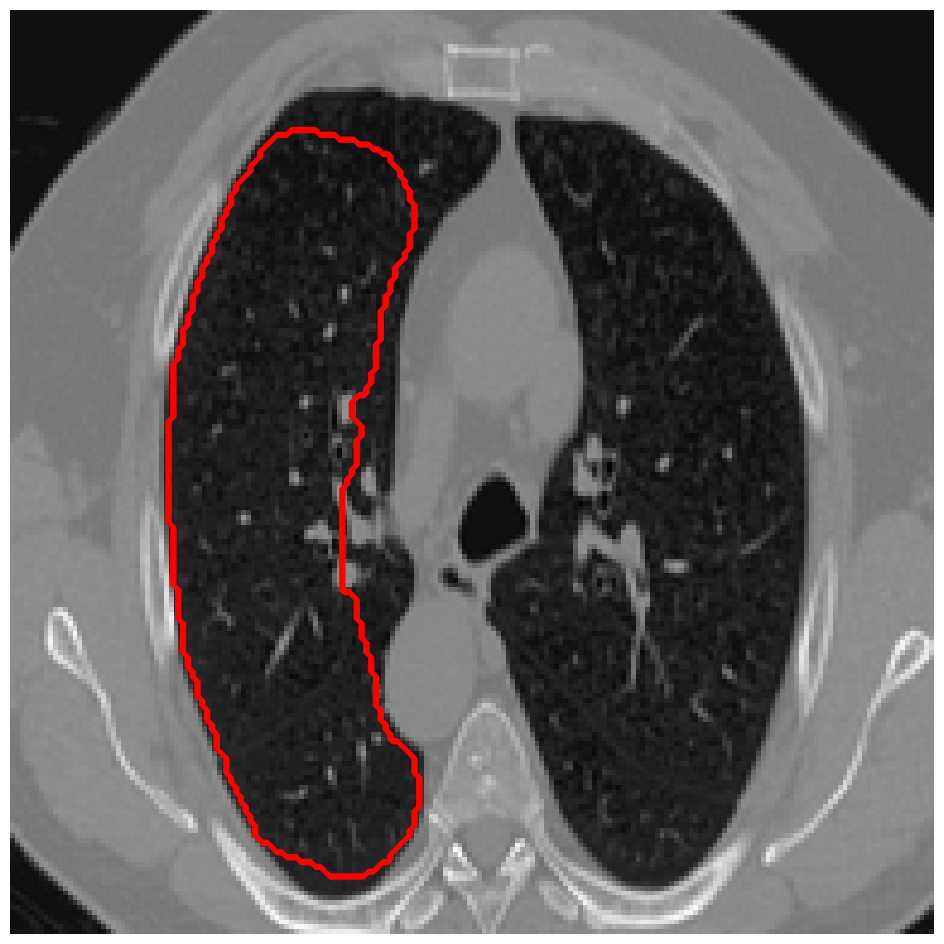}
    \caption{\textbf{M1}}
 \end{subfigure} 
   \begin{subfigure}[b]{0.16\textwidth}
    \includegraphics[width=\textwidth]{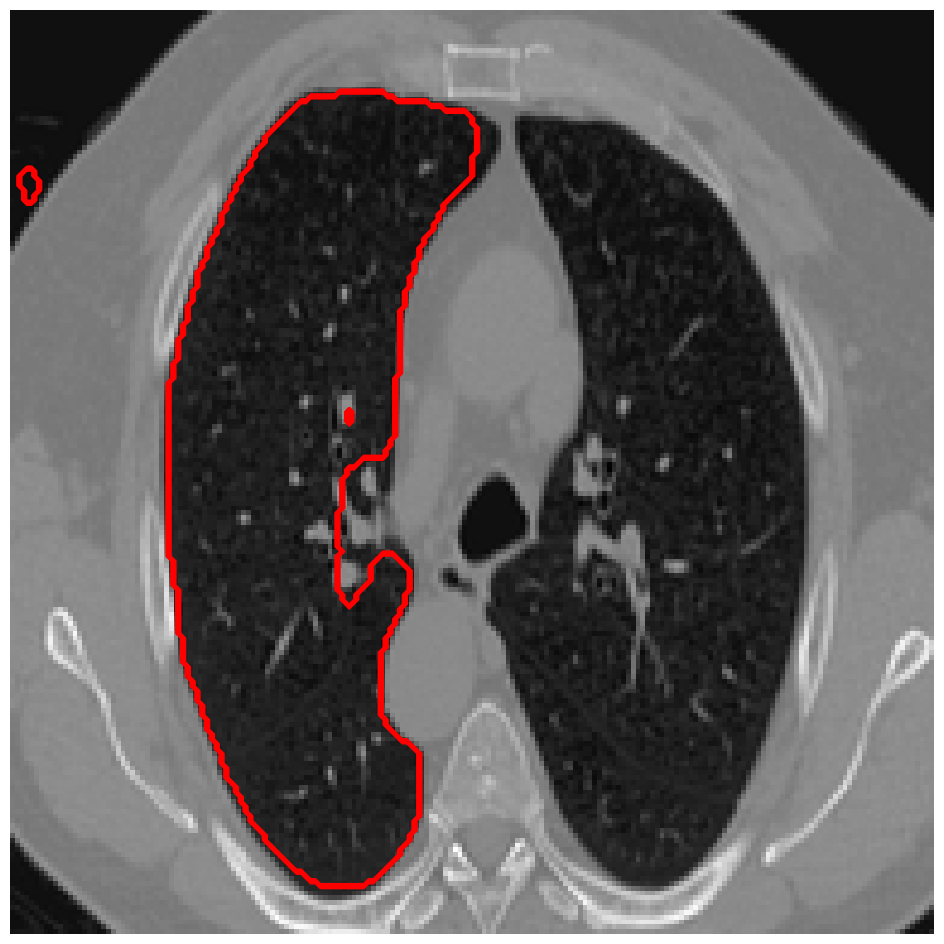}
    \caption{\textbf{M2}}
 \end{subfigure} \\
   \begin{subfigure}[b]{0.16\textwidth}
    \includegraphics[width=\textwidth]{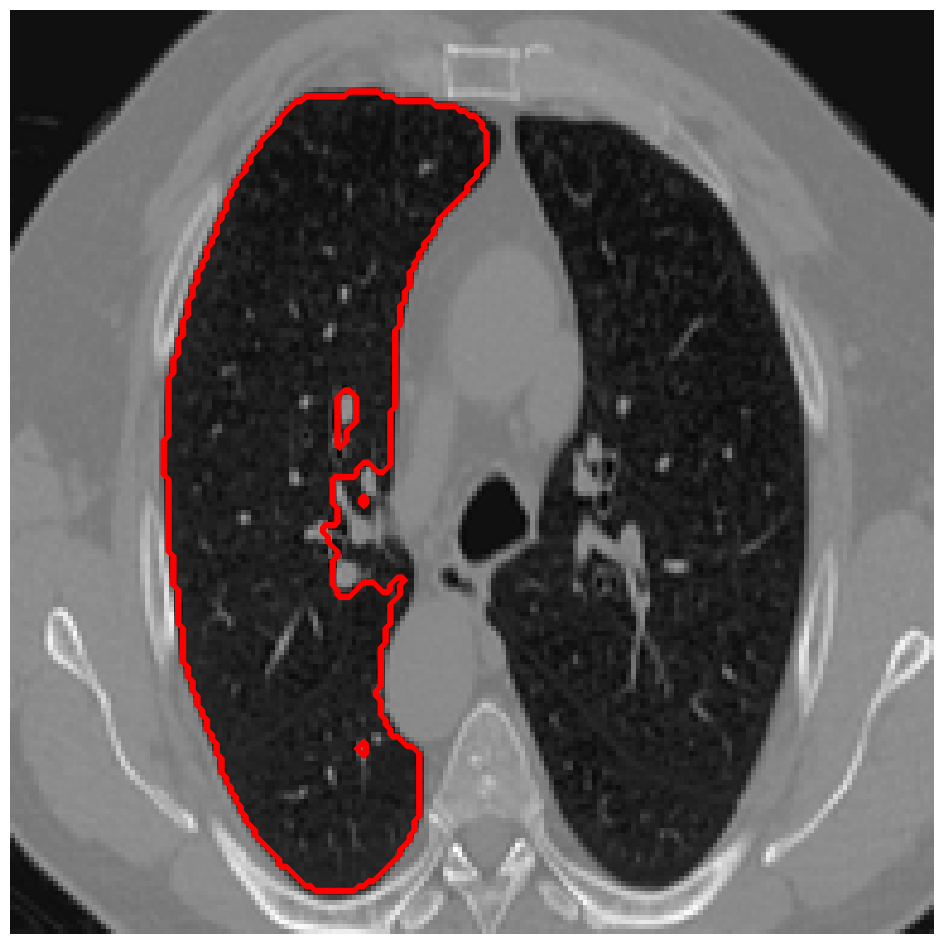}
    \caption{\textbf{M3}}
 \end{subfigure}
   \begin{subfigure}[b]{0.16\textwidth}
    \includegraphics[width=\textwidth]{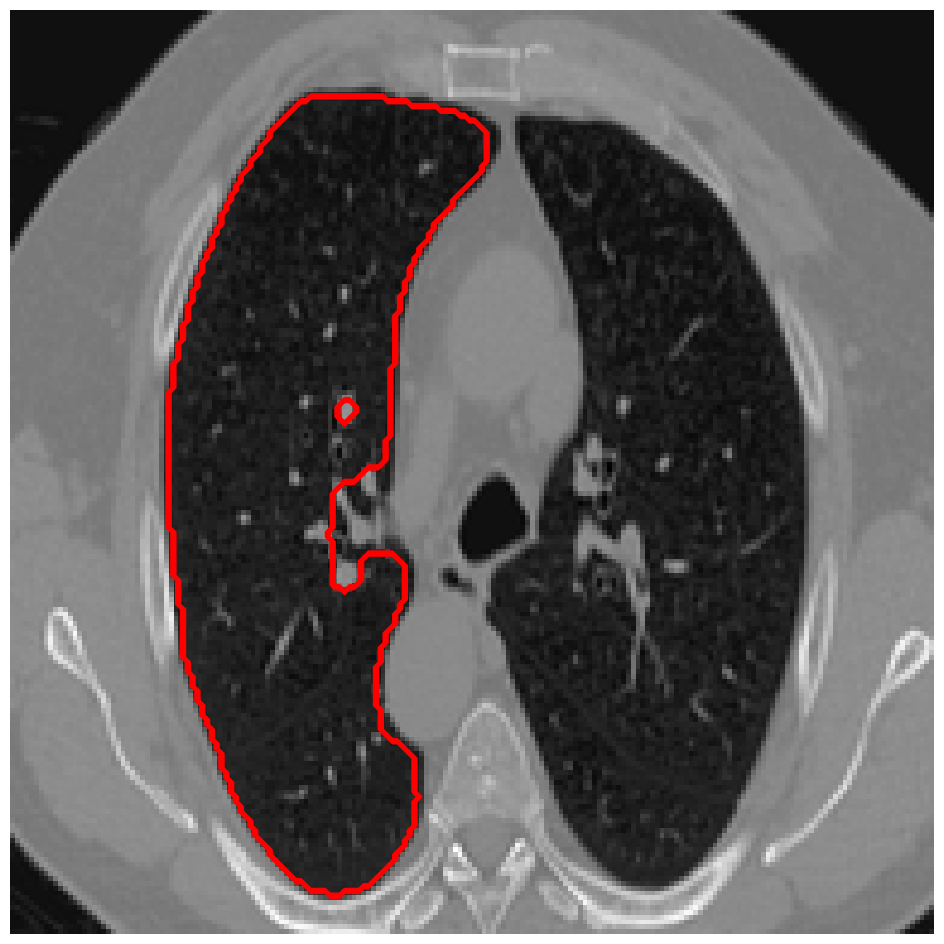}
    \caption{\textbf{M4}}
 \end{subfigure} 
 \begin{subfigure}[b]{0.16\textwidth}
    \includegraphics[width=\textwidth]{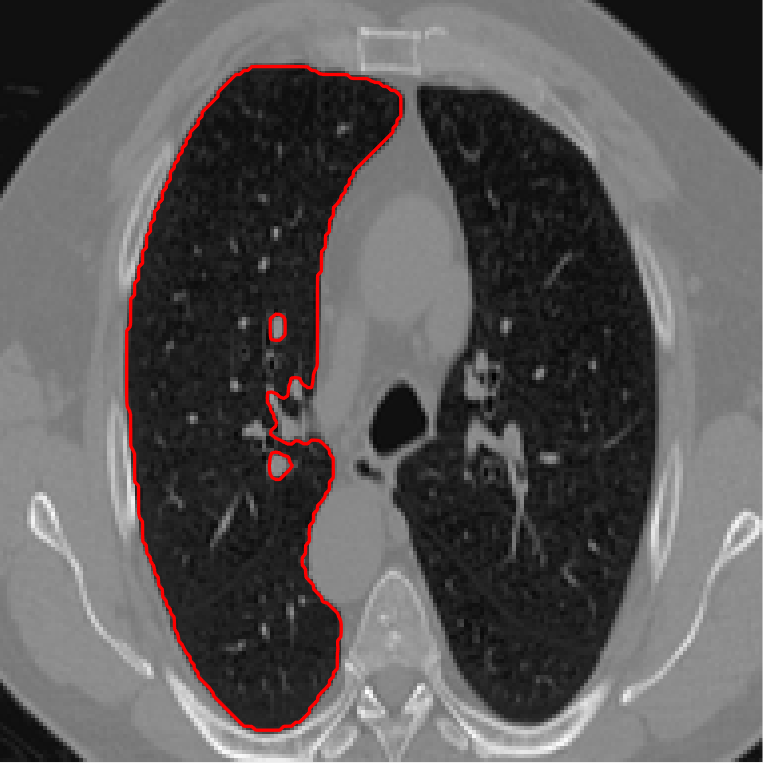}
    \caption{\textbf{TV-model}}
 \end{subfigure} 
 \begin{subfigure}[b]{0.16\textwidth}
    \includegraphics[width=\textwidth]{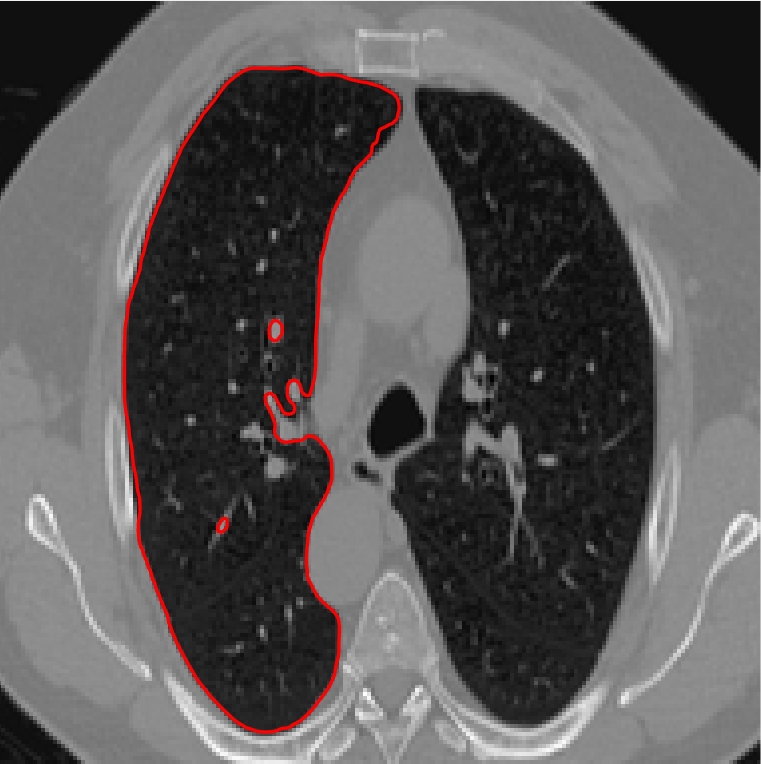}
    \caption{\scriptsize{\textbf{Elastica-model}}}
 \end{subfigure} 
 \begin{subfigure}[b]{0.16\textwidth}
    \includegraphics[width=\textwidth]{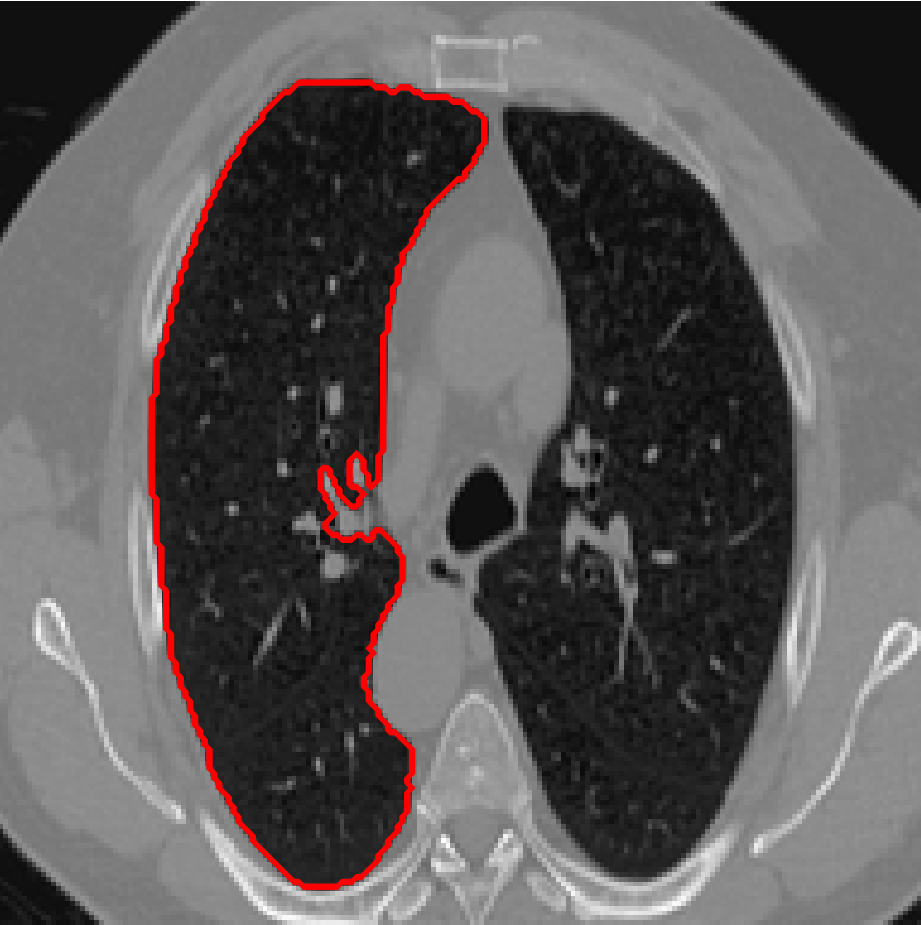}
    \caption{\scriptsize{\textbf{DIP-like-model}}}
 \end{subfigure} 
 \caption{A sample result on the Lung data. We display the input image with the user input $\mathcal{M}$, the ground truth (GT) and results from the four methods. Moreover, we show comparisons with the model \eqref{eq:RS-General} solved in  a variational framework with both Total Variation (TV) and Euler Elastica as explicit regularisation, as well as a comparison with the model solved in a Deep Image Prior framework.}
 \label{fig:Lung2}
\end{figure}

\begin{figure}[!htb]
\centering
 \begin{subfigure}[b]{0.16\textwidth}
    \includegraphics[width=\textwidth]{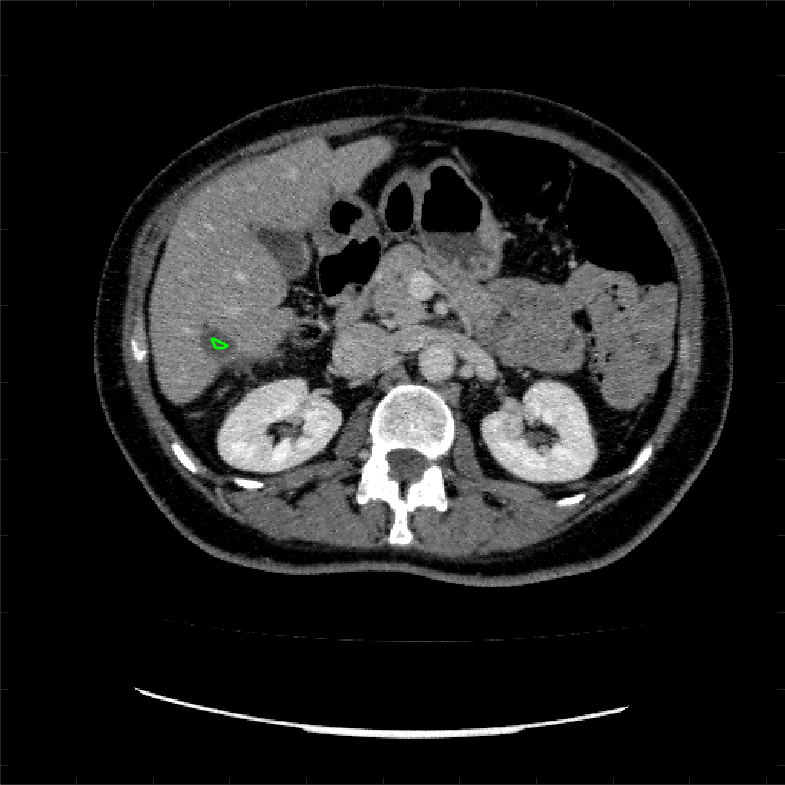}
    \caption{Image}
 \end{subfigure}
  \begin{subfigure}[b]{0.16\textwidth}
    \includegraphics[width=\textwidth]{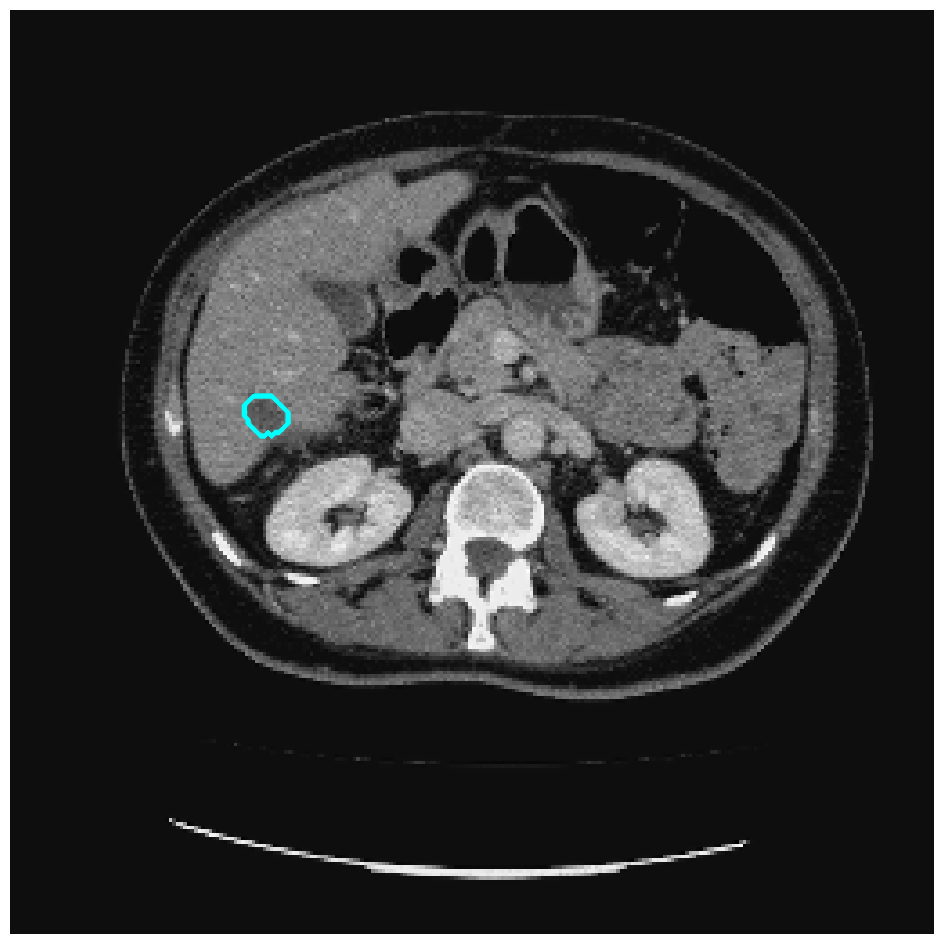}
    \caption{GT}
 \end{subfigure}
  \begin{subfigure}[b]{0.16\textwidth}
    \includegraphics[width=\textwidth]{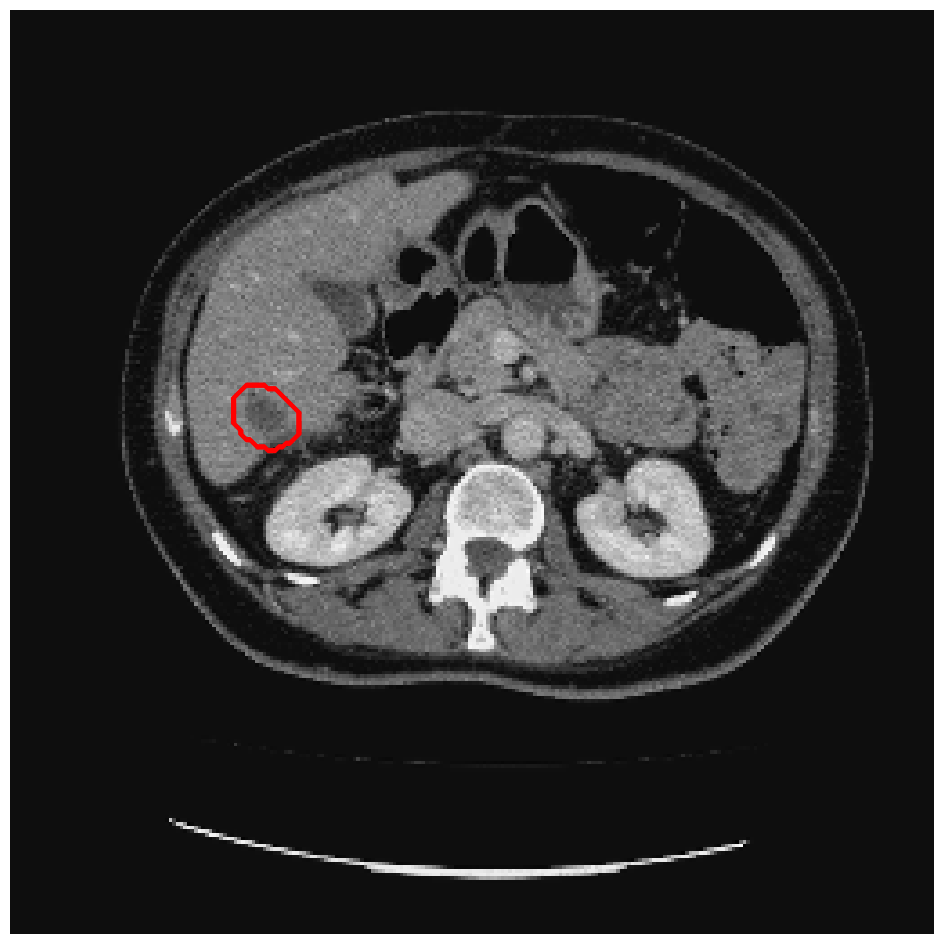}
    \caption{\textbf{M1}}
 \end{subfigure} 
   \begin{subfigure}[b]{0.16\textwidth}
    \includegraphics[width=\textwidth]{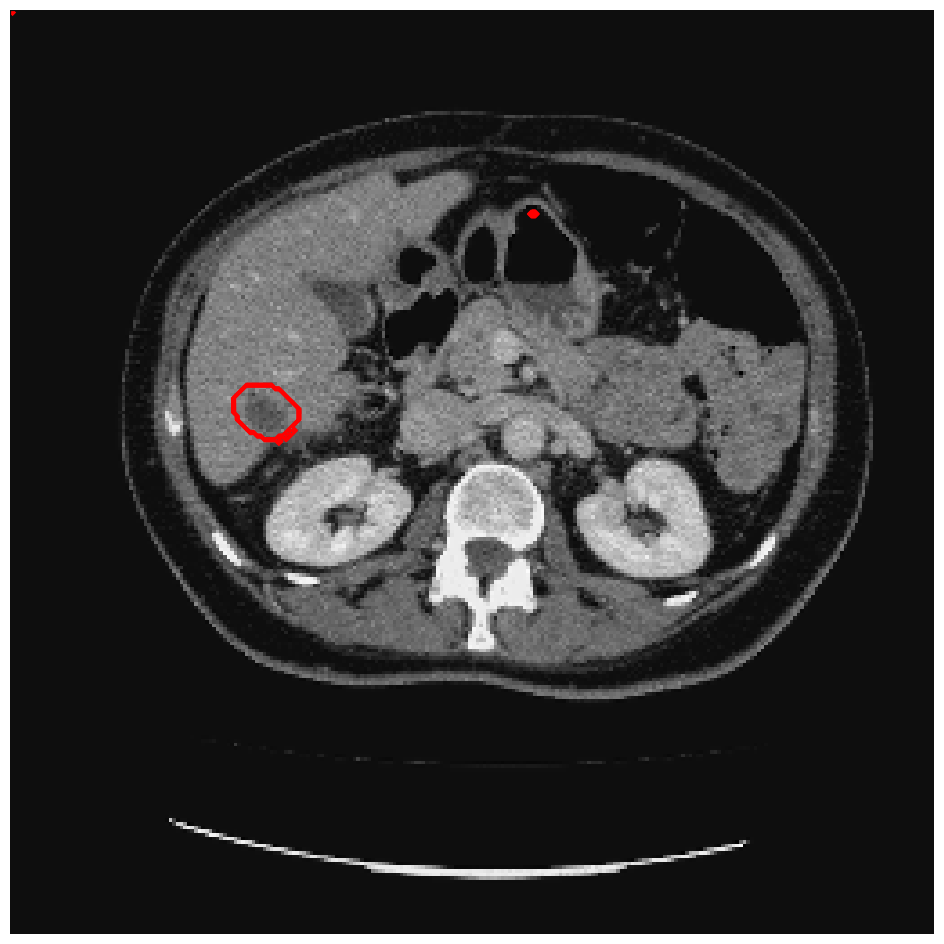}
    \caption{\textbf{M2}}
 \end{subfigure} \\
   \begin{subfigure}[b]{0.16\textwidth}
    \includegraphics[width=\textwidth]{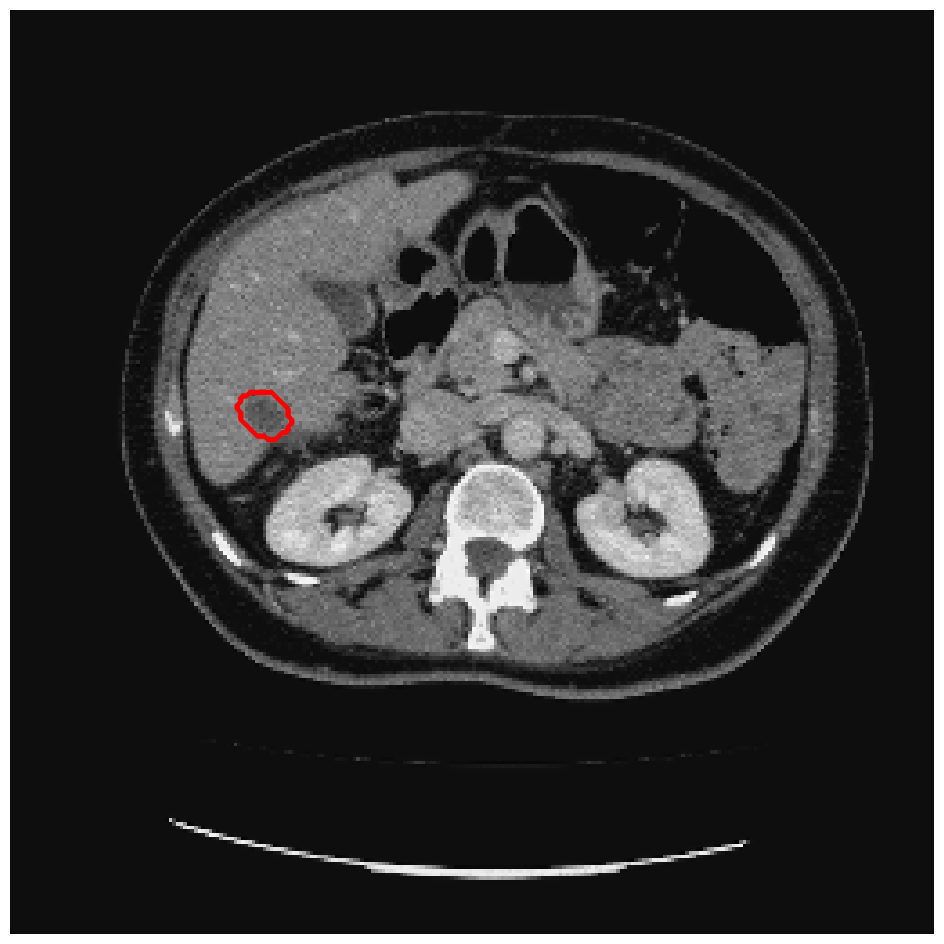}
    \caption{\textbf{M3}}
 \end{subfigure}
   \begin{subfigure}[b]{0.16\textwidth}
    \includegraphics[width=\textwidth]{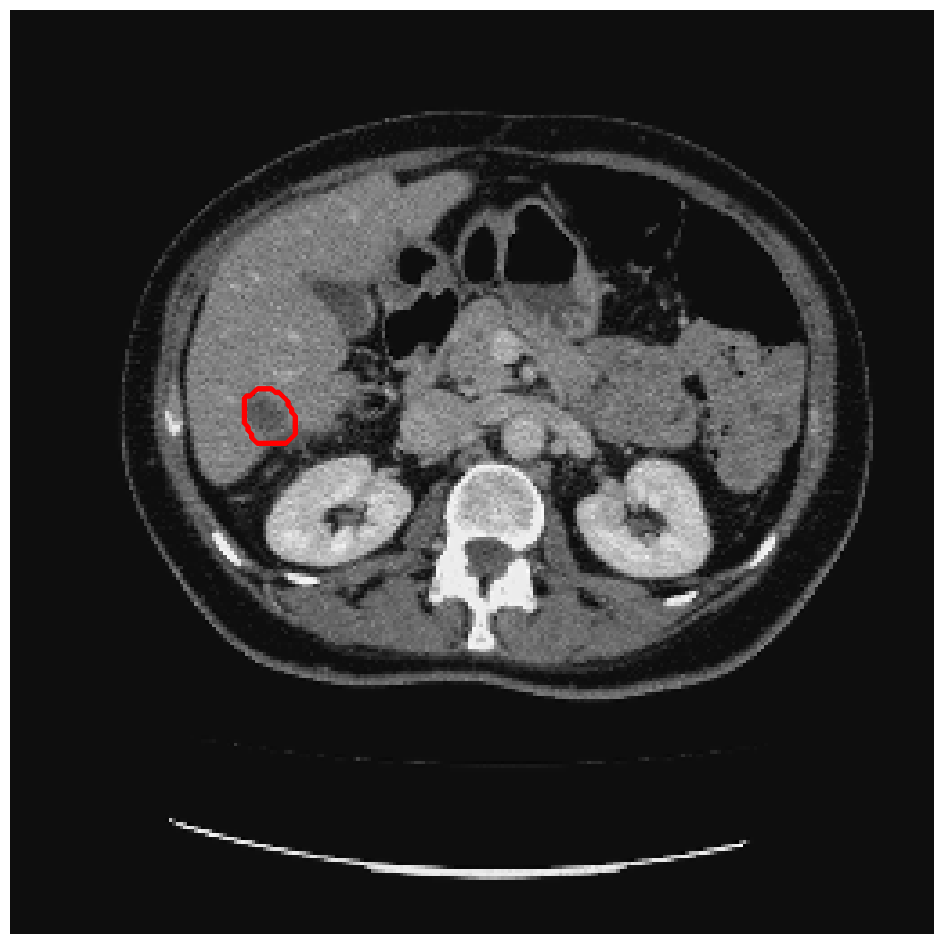}
    \caption{\textbf{M4}}
 \end{subfigure} 
 \begin{subfigure}[b]{0.16\textwidth}
    \includegraphics[width=\textwidth]{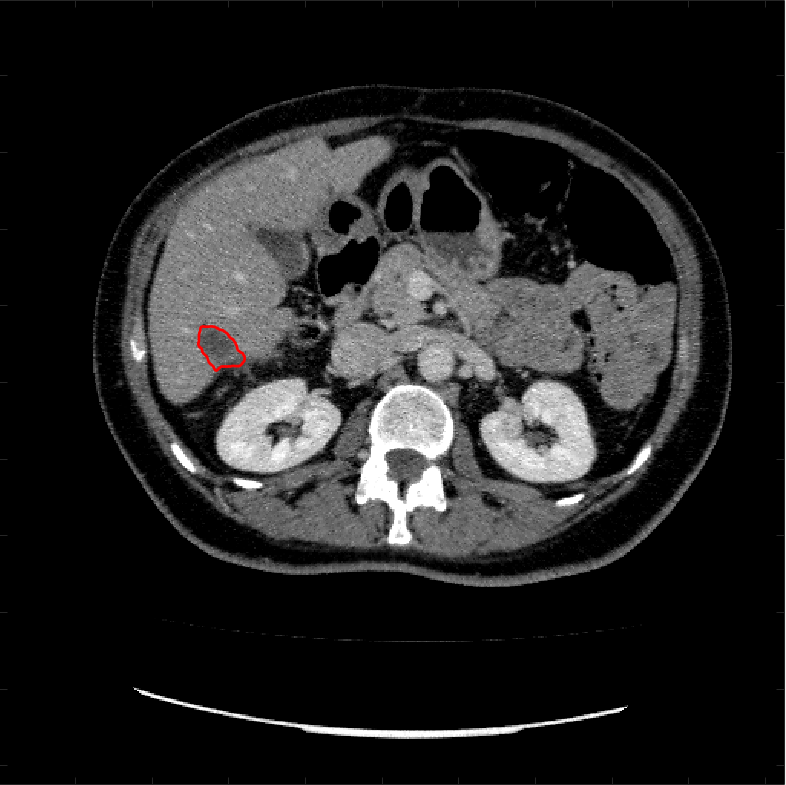}
    \caption{\textbf{TV-model}}
 \end{subfigure} 
 \begin{subfigure}[b]{0.16\textwidth}
    \includegraphics[width=\textwidth]{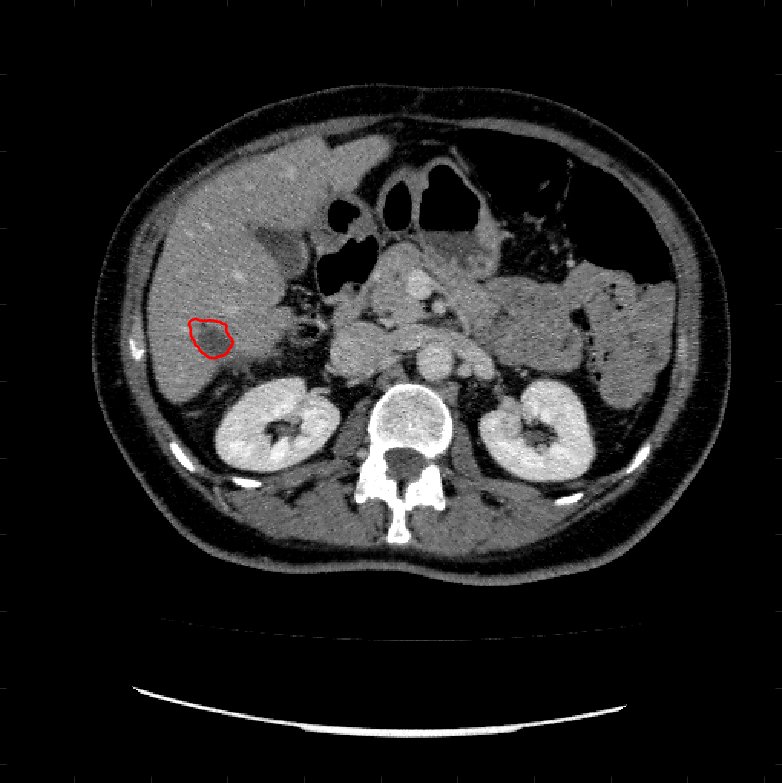}
    \caption{\scriptsize{\textbf{Elastica-model}}}
 \end{subfigure} 
 \begin{subfigure}[b]{0.16\textwidth}
    \includegraphics[width=\textwidth]{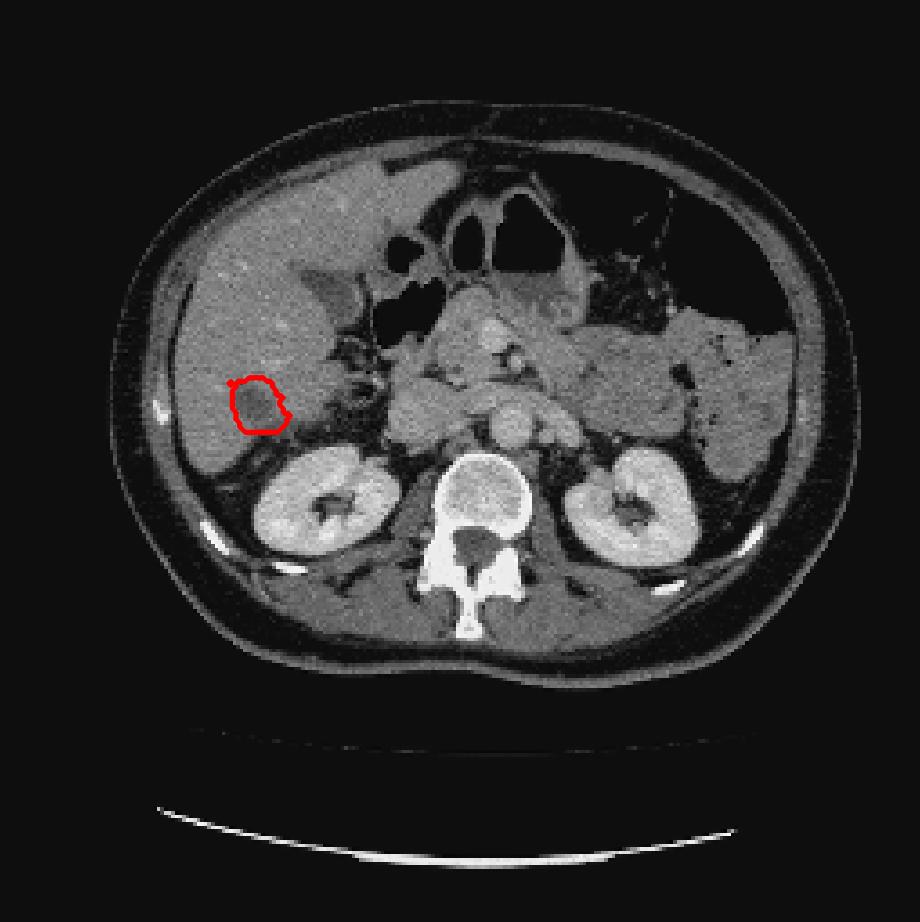}
    \caption{\scriptsize{\textbf{DIP-like-model}}}
 \end{subfigure} 
 \caption{A sample result on the Liver data. We display the input image with the user input $\mathcal{M}$, the ground truth (GT) and results from the four methods. Moreover, we show comparisons with the model \eqref{eq:RS-General} solved in  a variational framework with both Total Variation (TV) and Euler Elastica as explicit regularisation, as well as a comparison with the model solved in a Deep Image Prior framework.}
 \label{fig:Liver1}
\end{figure}

\begin{figure}[!htb]
\centering
 \begin{subfigure}[b]{0.16\textwidth}
    \includegraphics[width=\textwidth]{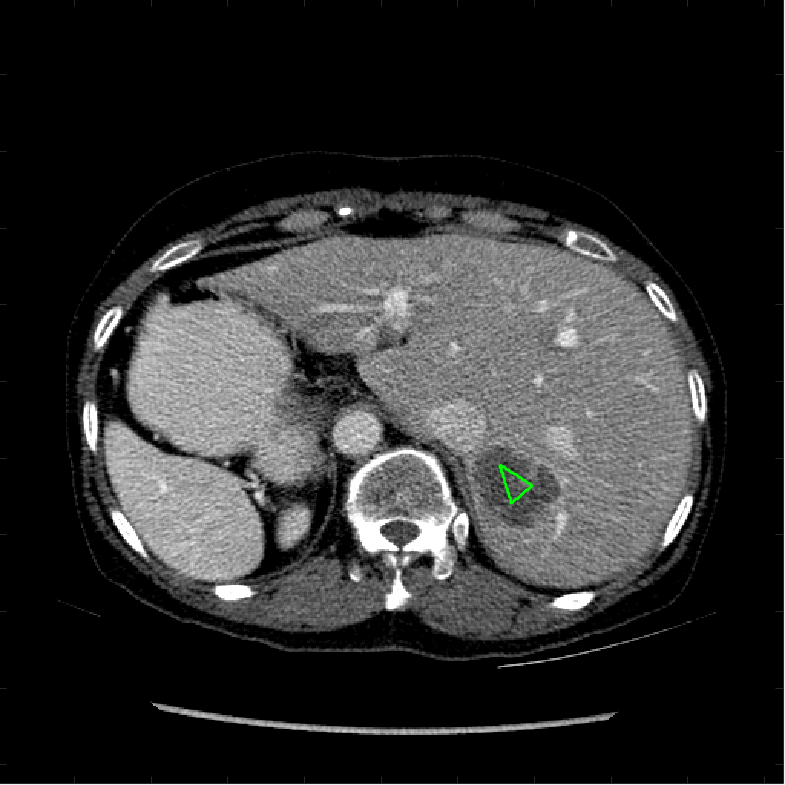}
    \caption{Image}
 \end{subfigure}
  \begin{subfigure}[b]{0.16\textwidth}
    \includegraphics[width=\textwidth]{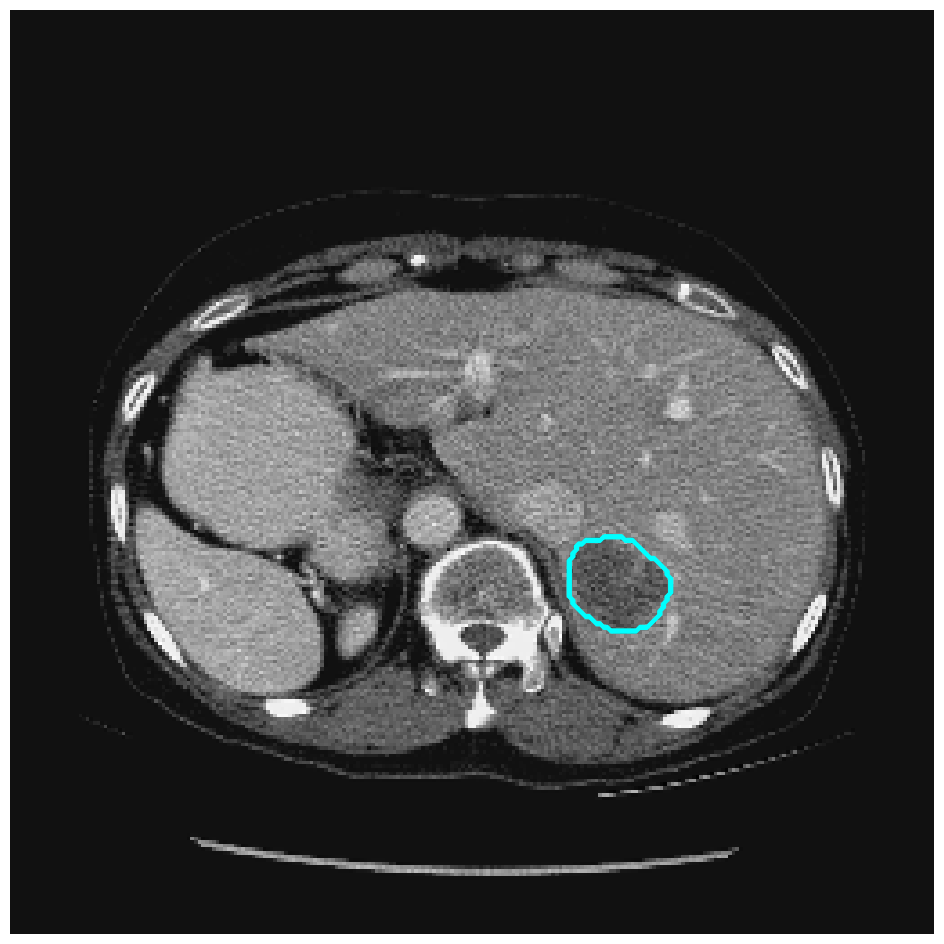}
    \caption{GT}
 \end{subfigure}
  \begin{subfigure}[b]{0.16\textwidth}
    \includegraphics[width=\textwidth]{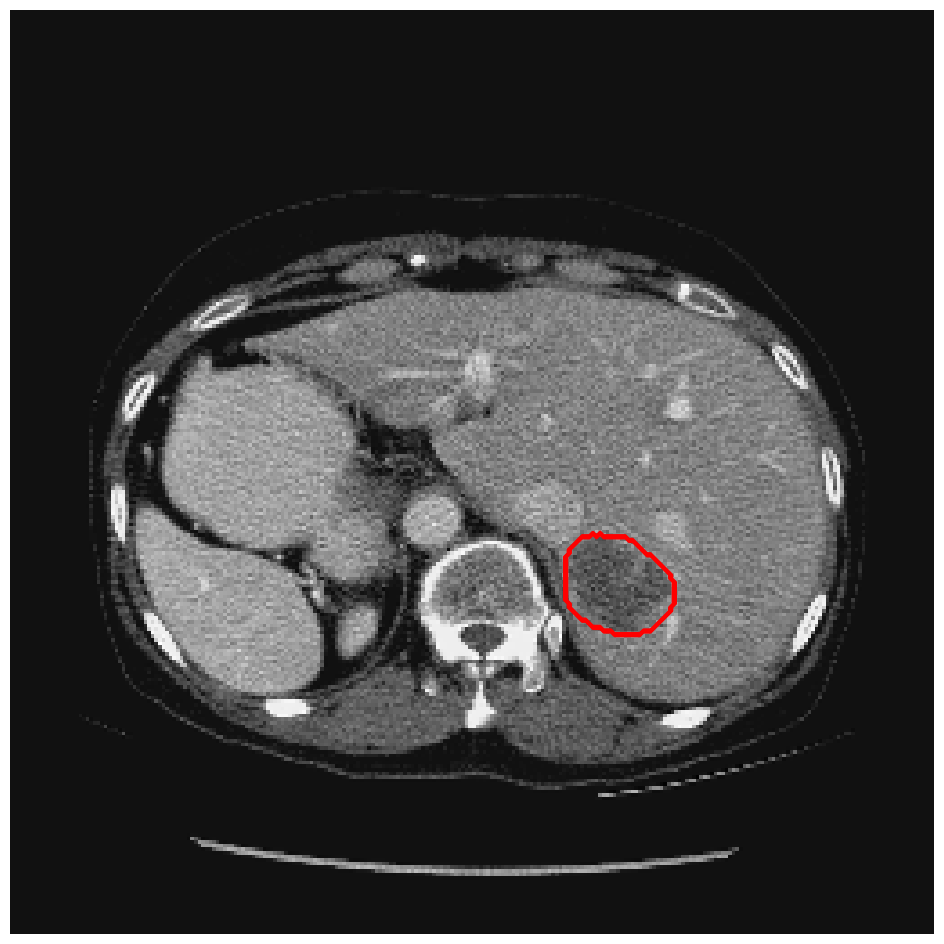}
    \caption{\textbf{M1}}
 \end{subfigure} 
   \begin{subfigure}[b]{0.16\textwidth}
    \includegraphics[width=\textwidth]{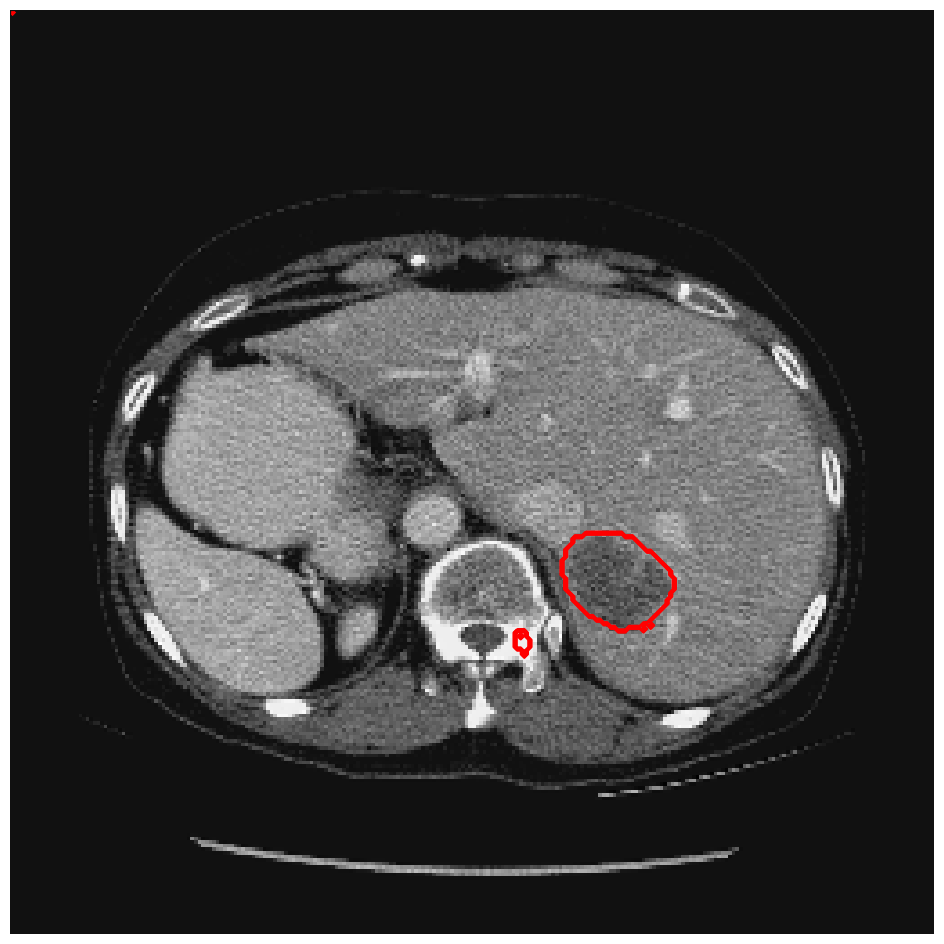}
    \caption{\textbf{M2}}
 \end{subfigure} \\
   \begin{subfigure}[b]{0.16\textwidth}
    \includegraphics[width=\textwidth]{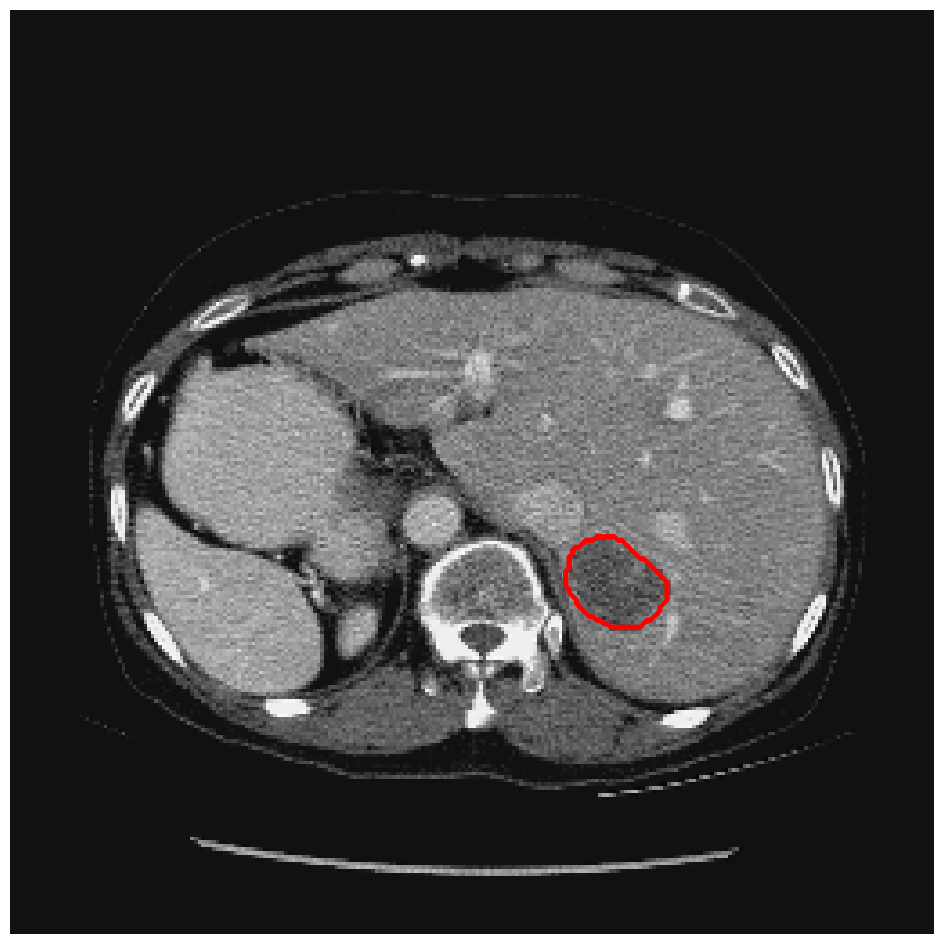}
    \caption{\textbf{M3}}
 \end{subfigure}
   \begin{subfigure}[b]{0.16\textwidth}
    \includegraphics[width=\textwidth]{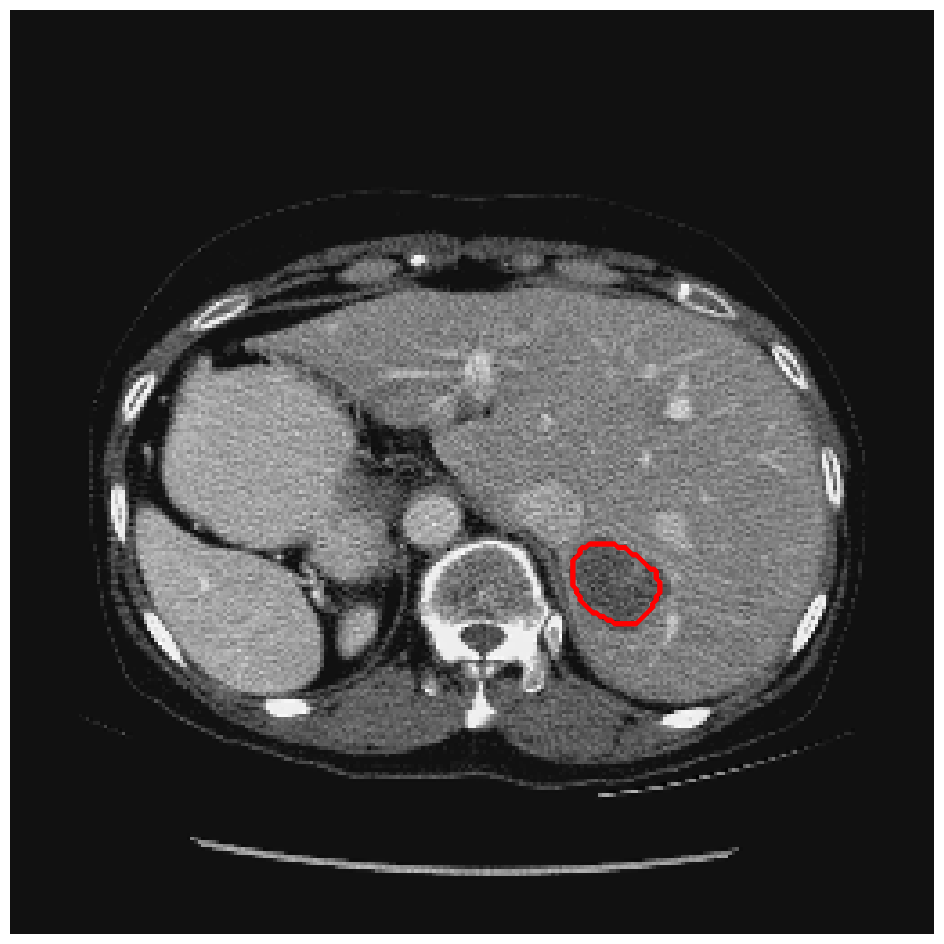}
    \caption{\textbf{M4}}
 \end{subfigure} 
 \begin{subfigure}[b]{0.16\textwidth}
    \includegraphics[width=\textwidth]{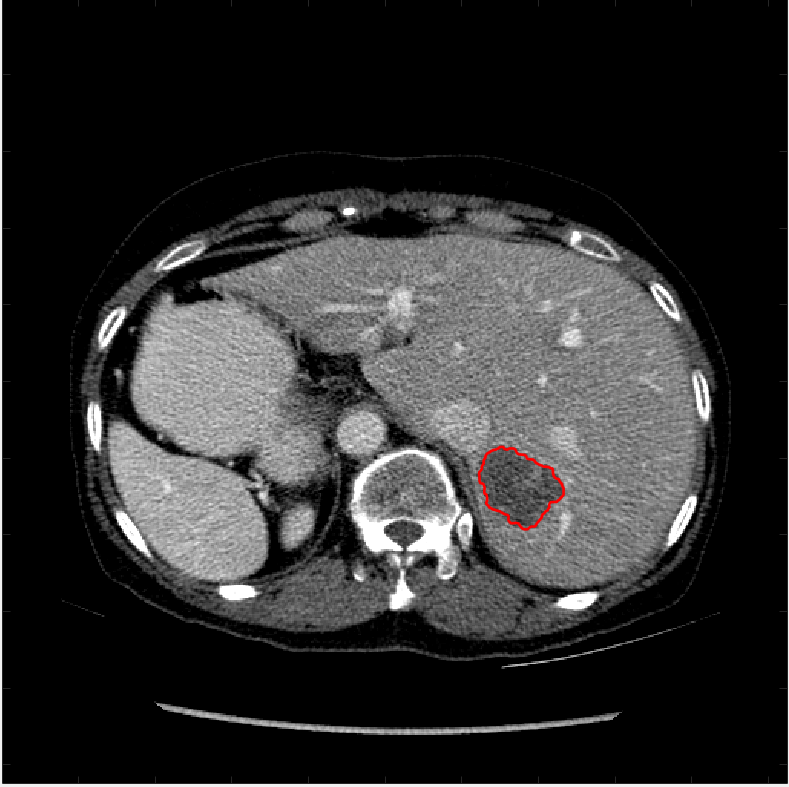}
    \caption{\textbf{TV-model}}
 \end{subfigure} 
 \begin{subfigure}[b]{0.16\textwidth}
    \includegraphics[width=\textwidth]{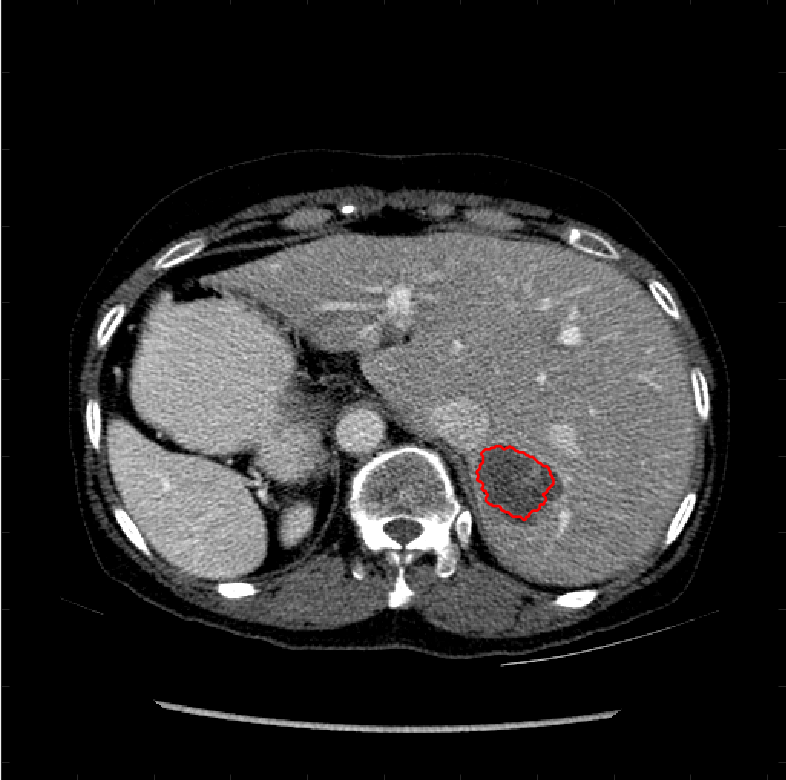}
    \caption{\scriptsize{\textbf{Elastica-model}}}
 \end{subfigure} 
 \begin{subfigure}[b]{0.16\textwidth}
    \includegraphics[width=\textwidth]{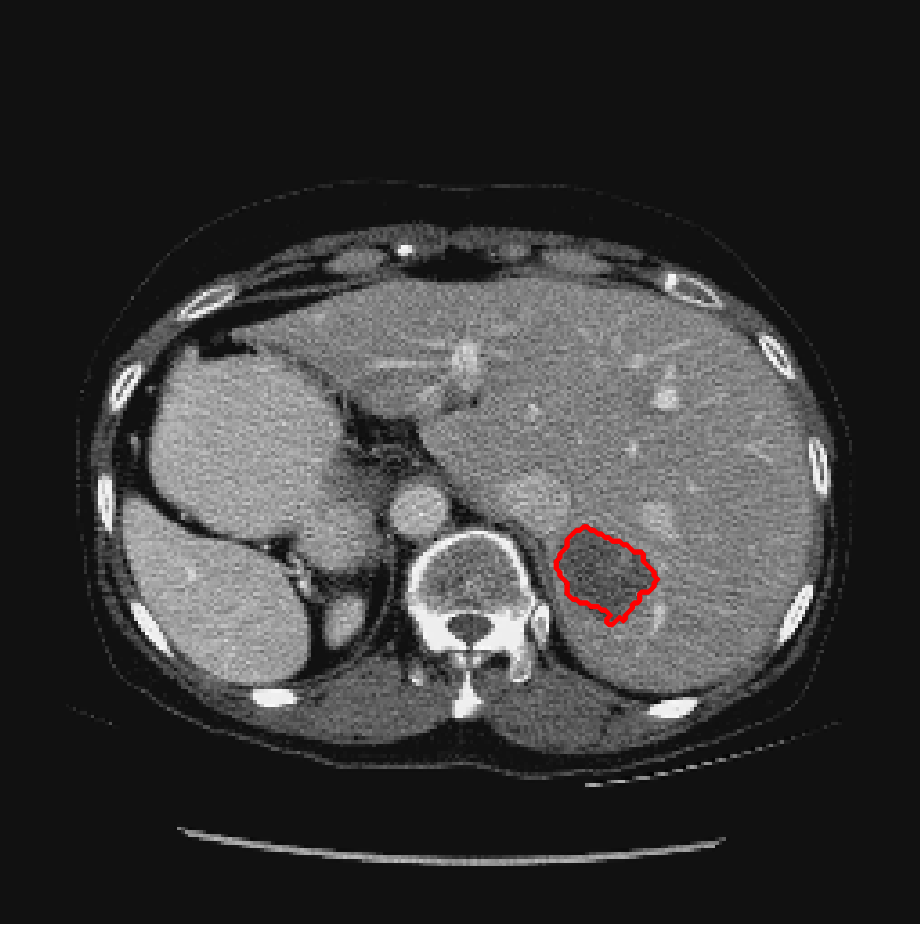}
    \caption{\scriptsize{\textbf{DIP-like-model}}}
 \end{subfigure} 
 \caption{A sample result on the Liver data. We display the input image with the user input $\mathcal{M}$, the ground truth (GT) and results from the four methods. Moreover, we show comparisons with the model \eqref{eq:RS-General} solved in  a variational framework with both Total Variation (TV) and Euler Elastica as explicit regularisation, as well as a comparison with the model solved in a Deep Image Prior framework.}
 \label{fig:Liver2}
\end{figure}

We conduct experiments on two datasets. The first selected data is images of the Lung from the Learn2Reg challenge dataset \footnote{Task 2: https://learn2reg.grand-challenge.org/Datasets/} \cite{LungDataset}. We selected images of the lung to demonstrate the advantages of our method as lung segmentation with variational methods with explicit regularisation can be difficult to achieve a smooth result when nodules are present. We chose $18$ images, one from each patient provided and used $2$ to train and $16$ to evaluate the performance. We aimed to segment the left lung by providing the some clicks on the region in order to provide the network with geometric information.

 \begin{table}[]
 \centering
\begin{tabular}{l|lllllll}
      & M1    & M2    & M3    & M4 & TV & Elastica & DIP   \\ \hline
DICE mean & 0.902 & 0.935 & 0.971 & 0.976 & 0.960 & 0.963 & 0.916 \\
DICE std  & 0.047 & 0.048 & 0.021 & 0.024 & 0.031 & 0.015 & 0.143 \\
JACCARD mean & 0.825 & 0.882 & 0.944 & 0.953 & 0.925 & 0.929 & 0.830 \\
JACCARD std & 0.077 & 0.082 & 0.038 & 0.044 & 0.052 & 0.027 & 0.168
\end{tabular}
\caption{Results for the $16$ test images of the CT Lung data.}
\label{tab:lung}
\end{table}

The second dataset we use is the Liver Tumour Segmentation Benchmark (LiTS) \footnote{https://competitions.codalab.org/competitions/17094} \cite{bilic2019liver}, focusing on only segmenting the lesions of the liver. We chose $29$ images randomly, and used $2$ images for training and $27$ to evaluate performance. Segmentation of the lesions was performed by providing the network with input inside the lesions for each image, so the network could be provided with geometric information.

\section{Results}
In this section we show results of the models discussed previously with the proposed approach. In order to show some quantitative results, we compute some common metrics to evaluate numerically the performance of the outputs of each method. We will display both the DICE coefficient and Jaccard coefficient. Given a segmentation result $\Sigma$ for an image  and an associated ground truth $GT$ manually annotated by an expert, the DICE coefficient is given by:
$$ DICE(\Sigma, GT) = \frac{2 | \Sigma \cap GT |}{| \Sigma | + | GT |}.$$
A coefficient of $1$ corresponds to a perfect result, whereas a value of $0$ is the opposite. Similarly, the Jaccard coefficient is defined as:
$$ JACCARD(\Sigma, GT) = \frac{| \Sigma \cap GT |}{| \Sigma | + |GT| - | \Sigma \cup GT|}.$$

In Figures \ref{fig:Lung1} and \ref{fig:Lung2} we show two select images from the test set for the Lung images to qualitatively show some results. To begin, we show some results of the model \eqref{eq:RobertsSpencer} solved with a Total Variation (TV) \cite{rudin1992nonlinear} regulariser, and with a Euler Elastica \cite{zhu2013image} regulariser. In addition, we show a result with the model \eqref{eq:RobertsSpencer} in a typical Deep Image Prior framework (i.e. training a network specifically for that image without the explicit regularisation in the loss function, employing early stopping). We see that the result from \textbf{M4} is an improvement over the TV and Elastica models, whereas results from the DIP example is comparable. However, the DIP example requires a new network trained specifically for the new image, whereas our result is acquired after training. In addition to the explicit regularisation comparison, we also display the results from the four methods (all of which are trained previously on $2$ images and used for prediction). Moreover some quantitative results are shown in table \ref{tab:lung}, which shows the mean DICE score and standard deviation on the 16 images in the test set.
 
Similarly we show some results in Figures \ref{fig:Liver1} and \ref{fig:Liver2} from the LiTS dataset of all the methods. Quantitative results of the $27$ images can be found in \ref{tab:lits}.
Clearly for both datasets, \textbf{M3} and \textbf{M4} using the proposed ideas outperform \textbf{M1} and \textbf{M2}, as well as methods using explicit regularisation (TV and Elastica), and the original DIP method.

In addition, we can also compared with \cite{burrows2020new} who also used the LiTS dataset. The unsupervised approach trained on over $1000$ images gave a mean DICE score of $0.671$ with a standard deviation of $0.305$ on $232$ validation images. Using our method $\textbf{M4}$ trained on just $2$ images give a mean DICE score of $0.753$ and standard deviation $0.213$.

 \begin{table}[]
 \centering
\begin{tabular}{l|lllllll}
     & M1    & M2    & M3    & M4  & TV & Elastica & DIP  \\ \hline
DICE mean & 0.841 & 0.767 & 0.879 & 0.868 & 0.862 & 0.864 & 0.840 \\
DICE std  & 0.137 & 0.224 & 0.071 & 0.080 & 0.090& 0.075 & 0.132 \\
JACCARD mean & 0.745 & 0.665 & 0.790 & 0.774 & 0.768 & 0.770 & 0.743  \\
JACCARD std & 0.174 & 0.241 & 0.112 & 0.116 & 0.129 & 0.122 & 0.165
\end{tabular}
\caption{Results for the $27$ test images of the LiTS liver tumour dataset.}
\label{tab:lits}
\end{table}

\section{Conclusion}

In this paper we have introduced a new method that leverages the implicit regularisation offered by the Deep Image Prior method, while retaining the traditional deep learning power of predicting unseen data. Our method needs very limited training data (in both experiments only two training images were used), and is able to be used for prediction on new images of a similar class.

We evaluate our proposed method using two baseline methods: one which is a typical variational method-like network with explicit regularisation, and one without. We find that the two methods that use our proposed approach with a deep image prior network working with a variational method-like network outperforms the two baseline methods on both datasets used.

\bibliography{reference.bib}
\bibliographystyle{spiebib}

\end{document}